\pdfoutput=1
\documentclass[11pt]{article}

\usepackage[preprint]{acl}

\usepackage{times}
\usepackage{latexsym}
\usepackage{pifont}
\newcommand{\cmark}{\ding{51}}%
\newcommand{\xmark}{\ding{55}}%
\usepackage[T1]{fontenc}
\usepackage[utf8]{inputenc}

\usepackage{microtype}

\usepackage{inconsolata}

\usepackage{graphicx}
\usepackage{bm}
\usepackage{subfigure}
\usepackage{amsmath}
\usepackage{amssymb}
\usepackage{multirow}
\usepackage{algorithm}
\usepackage{algpseudocode}

\allowdisplaybreaks
\usepackage{kotex}

\title{PromptKD: Distilling Student-Friendly Knowledge for Generative Language Models via Prompt Tuning}

\author{Gyeongman Kim\textnormal{\textsuperscript{1}} \quad Doohyuk Jang\textnormal{\textsuperscript{1}} \quad Eunho Yang\textnormal{\textsuperscript{1,2}} \vspace{0.03in}\\
  \textsuperscript{1}Korea Advanced Institute of Science and Technology (KAIST), South Korea \\ \textsuperscript{2}AITRICS, South Korea\\
  \texttt{\{gmkim, jadohu, eunhoy\}@kaist.ac.kr}
  }
\begin{document}
\maketitle
\begin{abstract}
Recent advancements in large language models (LLMs) have raised concerns about inference costs, increasing the need for research into model compression. 
%While knowledge distillation (KD) is a prominent method for this, most existing studies focus on classification models, making it challenging to apply to generative language models like LLMs. Additionally, there have been observations that adapting the teacher to the student's capacity and distilling student-friendly knowledge can improve performance, but such ideas remain unexplored in generative language models. 
While knowledge distillation (KD) is a prominent method for this, research on KD for generative language models like LLMs is relatively sparse, and the approach of distilling student-friendly knowledge, which has shown promising performance in KD for classification models, remains unexplored in generative language models. 
To explore this approach, we propose PromptKD, a simple yet effective method that utilizes prompt tuning - for the first time in KD - to enable generative language models to transfer student-friendly knowledge. 
Unlike previous works in classification that require fine-tuning the entire teacher model for extracting student-friendly knowledge, PromptKD achieves similar effects by adding a small number of prompt tokens and tuning only the prompt with student guidance. 
% Therefore, PromptKD is a scalable approach applicable even to large models like LLMs. 
Extensive experiments on instruction-following datasets show that PromptKD achieves state-of-the-art performance while adding only 0.0007\% of the teacher's parameters as prompts. Further analysis suggests that distilling student-friendly knowledge alleviates exposure bias effectively throughout the entire training process, leading to performance enhancements.\footnote{Project page: \href{https://promptkd.github.io}{https://promptkd.github.io}}
% 최근 LLM이 빠른 속도로 발전하면서 inference cost에 대한 문제도 함께 대두되어 모델을 경량화하는 연구에 대한 필요성이 증대되고 있다. 경량화 방식 중 prominent한 방법으론 KD가 있지만, 기존 연구는 대부분 classification model을 대상으로하여 LLM과 같은 generative language model에 적용하기 어려운 상황이다. 또한, teacher가 student의 capacity를 aware한 채 변형되어 student-friendly knowledge를 distill한다면 성능이 좋아지는 observation들도 있었는데, 이러한 아이디어 역시 generative language model에서는 아직 explore되지 않았다. 이를 밝혀내기 위해, 우리는 knowledge distillation 연구에서 처음으로 prompt를 활용하여 generative language model이 student-friendly knoweldge를 전수할 수 있도록 하는 simple yet effective한 promptkd를 제안한다. PropmtKD는 LLM과 같은 큰 모델에도 쉽게 적용가능한 scalable method로, 기존 연구처럼 student-friendly knowledge를 추출하기 위해 teacher 전체를 변형시키지 않고, 아주 적은 수의 prompt token을 prepend하여 student의 guidance를 통해 prompt만 tuning함으로써 teacher 전체를 fine-tuning한 것과 같은 효과를 내었기 때문이다. GPT-2 family의 model을 사용하여 instruction-following dataset에 대해 PromptKD를 평가한 결과, teacher parameter의 0.000%만큼을 prompt로 추가하면서 평균 ~ 성능 증가를 이끌어 sota 성능을 달성했다. Further analysis를 통해 추측했을떄, student-friendly knowledge를 전수하는 것이 학습 초반에 exposure bias를 빠르게 해소하기 때문에 성능이 증가한 것으로 생각된다.
\end{abstract}

\section{Introduction}

\begin{figure}[!t]
\centering
\includegraphics[width=\linewidth]{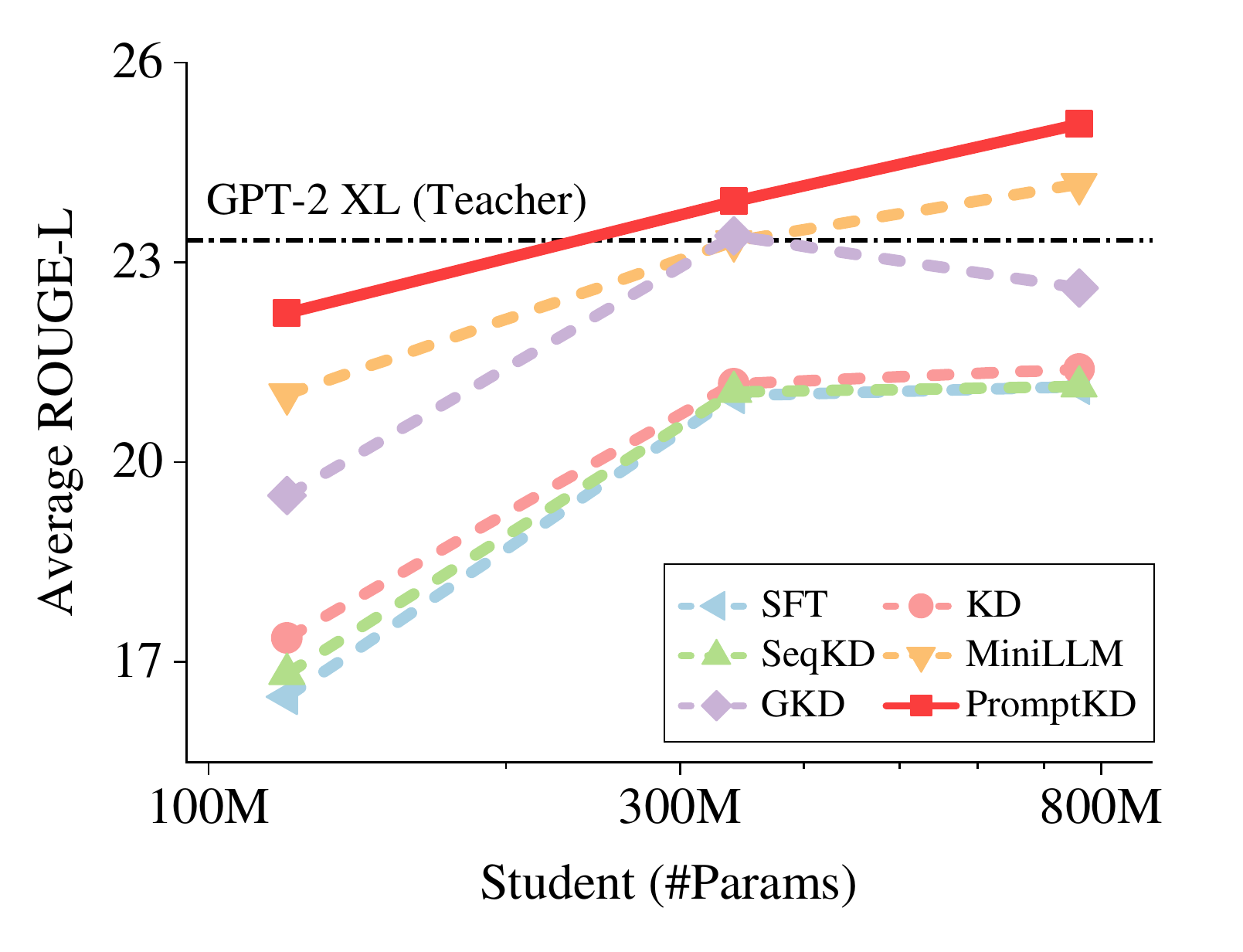}
\caption{Comparison of instruction-following performance of KD methods using the GPT-2 model family. Owing to the student-friendly knowledge, our PromptKD outperforms others with only an additional 11K parameters. Dashed reference line represents the performance of the teacher model.}
%\textbf{Average ROUGE-L scores} Each KD method conducted with GPT-2 XL as a teacher and GPT-2 small(125M), GPT-2 medium(340M), GPT-2 large(760M) as the students.The performance of the teacher model is represented by a dashed reference line.
\label{fig:intro_figure}
\end{figure}

With the massive improvement of generative language models, such as the emerging abilities~\citep{wei2022emergent} observed in large language models (LLMs), there is a growing need for model compression research to efficiently deploy models in various tasks~\citep{touvron2023llama,alpaca}. However, among notable compression methods such as knowledge distillation (KD; \citealp{hinton2015distilling,kim-rush-2016-sequence,minillm}), pruning~\citep{ma2023llmpruner}, and quantization~\citep{tao-etal-2022-compression}, KD has not been successfully applied to generative language models.
%applying KD to generative language models still remains challenging.
% LLM의 emerging ability와 같은 observation과 함께 generative language model이 다양한 task에서 빠르게 발전하면서, 많은 서비스에서 model을 효율적으로 deploy하기 위해 경량화 연구의 필요성도 꾸준히 커지고 있다 (sLLM). generative langauge model의 잘 알려진 경량화 방법으로는 knowledge disitllation, pruning, 그리고 quantization이 있으나, 아직 knowledge distillation을 이러한 모델에 적용하기에는 어려운 상황이다.

%The primary reason is that most KD methods are devised with classification tasks, dealing with classification models like BERT~\citep{devlin-etal-2019-bert}, making them unsuitable for application to generative language models.
Since most KD methods are devised with models like BERT~\citep{devlin-etal-2019-bert} for classification tasks, the challenge arises when attempting to directly apply these KD methods to generative language models, which have different architectures and are designed for tasks other than classification.
While there have been some methods proposed for generative language models, such as Supervised KD~\citep{sanh2019distilbert} or SeqKD~\citep{kim-rush-2016-sequence}, they tend to be naive approaches. Even recently proposed works~\citep{gkd,minillm}, like previous research, have focused on distribution discrepancy metrics or pseudo-targets. Therefore, despite the rapid advancement of LLMs in recent times, the drawback is that they are not designed with the extension to LLMs in mind.
%Even recently proposed works~\citep{gkd,minillm} do not focus on the large size of the teacher model, as they often concentrate on aspects such as distribution discrepancy metrics or pseudo-targets.
% 기본적인 이유는 대부분의 방법론들이 BERT와 같은 classification model을 다루면서 classification task를 목적으로 고안되었기 때문이다. 물론 LLM과 같은 generative langauge model을 위해 제안된 방법론도 몇가지 있지만 wordKD나 SeqKD처럼 naive한 방식이거나, 최근에 제안되었더라도 distribution discrepancy metric이나 Pseudo-Target에 집중하는 등 teacher의 크기가 큰 것을 신경쓰며 고안된 방법론이 아니란 점이 있다.

Moreover, attempts to distill student-friendly knowledge in a generative language model have yet to be explored. Recent KD studies~\citep{yang-etal-2022-sparse,park2021learning,zhou-etal-2022-bert} for classification tasks aim to distill such knowledge. This idea emerges because previous works extract knowledge from fixed teacher without knowing the student's capacity, and the observation~\citep{cho2019efficacy} that larger teacher models do not necessarily improve student performance. However, there hasn't been any exploration of applying these ideas to generative language models. Since the capacity gap between teacher and student persists in KD for generative language models, it is reasonable to expect that distilling student-friendly knowledge would be beneficial.
% 게다가, 아직 student-friendly knowledge를 전수하려는 시도가 explore되지 않은 상황이다. 최근 classification model을 위한 KD 연구들은 student-friendly한 knoweldge를 distill하고자 하는데, 이는 teacher가 student의 capacity를 모르는 상태에서 고정된 채 knowledge가 추출된다는 점과 teacher의 크기가 클수록 student의 성능이 좋아지는 것도 아니라는 observation 때문이다. 그러나, generative language model에 이러한 아이디어를 적용한 방법은 아직 없다. Generative language model의 KD에서도 teacher와 student 사이의 capacity gap은 여전히 존재하는 만큼, student-friendly knowledge를 전수하는 것이 도움이 될 것으로 기대할 수 있다.

To address this issues, we propose PromptKD, which utilizes prompts in generative language models to distill student-friendly knowledge. Extracting student-friendly knowledge from the teacher requires modifying the teacher, as in previous studies~\citep{ren-etal-2023-tailoring, zhou-etal-2022-bert}. However, modifying a large teacher model can incur significant computational costs. PromptKD addresses this concern by exploiting prompt tuning. By appending prompt tokens to the beginning of the input, we can efficiently fine-tune the teacher model with notably fewer parameters. While there are other parameter-efficient fine-tuning methods such as prefix-tuning~\citep{li-liang-2021-prefix} and LoRA~\citep{hu2022lora}, they suffer from the disadvantage that the number of parameters to be trained increases proportionally with the number of layers. Moreover, there is an observation~\citep{lester-etal-2021-power} that prompt tuning shows similar performance to full-parameter fine-tuning as the model size increases, making prompt tuning a more reasonable choice. PromptKD learns prompts that stimulate the teacher to distill student-friendly knowledge with guidance from the student. Additionally, it employs regularization loss during the early stages of training to prevent significant divergence from the original teacher when appending prompts, ensuring stable training.
% 이러한 아이디어를 explore하기 위해, 우리는 generative language model에서 prompt를 활용하여 student-friendly knowledge를 distill하는 PromptKD를 제안한다. 우선 teacher로부터 student-friendly knowledge를 추출하기 위해서는 기존 연구들처럼 teacher를 fine-tuning해야 하는데, teacher의 크기가 크다면 많은 computational cost가 필요하게 된다. PromptKD는 이 우려를 prompt tuning을 통해 해결한다. input 단에서 prompt를 활용하게되면 parameter-efficient하게 teacher model을 fine-tuning할 수 있다는 장점이 있다. 비록 다른 Parameter efficient fine-tuning 방법론(Prefix-tuning, LoRA)도 있지만, 모델의 크기(layer)에 따라 학습해야하는 parameter가 비례하여 증가한다는 단점이 있다. 또한, model의 크기가 커질수록 prompt tuning이 full fine-tuning과 유사한 성능을 보인다는 observation이 있어 prompt tuning은 더욱 합리적인 choice이다. PropmtKD는 student의 guidance을 통해 teacher로부터 student-friendly knowledge를 추출할 수 있도록 stimulate하는 prompt를 배워가며, prompt를 활용하는 학습 초반 과정에서 기존의 teacher model과 크게 멀어져 학습이 불안정해지는 것을 방지하는 regularization loss도 활용한다.

For evaluation, we measure the instruction-following performance~\citep{ouyang2022training}, aiming to cover a variety of tasks that generative language models can perform. Compared to the existing baseline, PromptKD achieves state-of-the-art performance by adding prompt parameters equivalent to only 0.0007\% of the teacher parameters, as depicted in Figure~\ref{fig:intro_figure}. Additionally, the analysis of exposure bias suggests that remarkable alleviation of exposure bias through student-friendly knowledge is likely the cause of performance improvement. Lastly, we explore the student-friendly knowledge in PromptKD and confirm the necessity of regularization loss and the importance of prompt initialization through ablation studies.
% 우리는 generative language model이 수행하는 다양한 NLP task를 커버할 수 있는 instruction-following performance를 평가하였으며, 기존 baseline과 비교해서 only teacher parameter의 0.000%만큼의 parameter 증가와 함께 sota 성능을 이끌어 냈다. 또한 student의 exposure bias를 측정한 결과, 학습 초반에 student-friendly knowledge를 통해 빠르게 exposure bias를 해소한 것이 성능 증가의 원인으로 예상된다. 마지막으로 ablation study를 통해 regualrization loss의 필요성과, prompt initialization의 중요성을 확인하였다.
% 마지막으로 student-friendly knowledge에 대해 explore하면서, ablation study를 통해 regualrization loss의 필요성과, prompt initialization의 중요성을 확인하였다.

To summarize, our contribution is four-fold:
\begin{itemize}
\item We investigate the effect of student-friendly knowledge, which has not been previously explored in knowledge distillation (KD) for generation tasks.
%\vspace{-0.1in}
\item We propose PromptKD, the first usage of prompt tuning in KD, enabling memory-efficient extraction of student-friendly knowledge from teacher.
%\vspace{-0.1in}
\item Through extensive experiments on 5 instruction-following datasets, PromptKD achieves state-of-the-art performance.
%\vspace{-0.1in}
\item We suggest that the superiority of PromptKD lies in its ability to fully mitigate exposure bias in the training phase.
\end{itemize} 
% 우리 연구의 contribution을 정리하면 다음과 같은데, - KD for generation task에서 밝혀지지 않았던 student-friendly knoweldge의 효과를 investigate한다. -knowledge distillation에서 처음으로 prompt tuning을 사용함으로써, memory-efficient하게 student-friendly knowledge를 추출하는 PromptKD를 제안한다. - 5개의 instruction-following dataset에 대한 extensive experiments 결과 SOTA 성능을 달성했다. - exposure bias에 대한 분석을 통해 성능 증가의 원인을 제안한다.

\section{Related Work}

\paragraph{KD for text classification}

Knowledge distillation (KD)~\citep{hinton2015distilling} is a model compression technique where the knowledge of a teacher model is transferred to improve the performance of a student model. Most KD research has been focused on text classification tasks. It has evolved from simple approaches~\citep{song2020lightpaff} that match the class distributions between teacher and student to more complex methods~\citep{jiao-etal-2020-tinybert,sun-etal-2019-patient,wang2020minilm,park-etal-2021-distilling} that involve matching hidden states or attention matrices between models. Recently, concerns have been raised about the observation~\citep{cho2019efficacy} that larger teacher models do not necessarily produce better students and the issue of teachers distilling knowledge while being unaware of the student's capacity. To address this, \citet{park2021learning,zhou-etal-2022-bert,ren-etal-2023-tailoring} transfer student-friendly knowledge, which requires the teacher to transform during the distillation process, influenced by specific objectives aimed at benefiting the student. Additionally, focusing on the capacity gap between the teacher and student during training, \citet{yang-etal-2022-sparse} proposes gradually pruning the teacher, while \citet{liang2023homodistil} suggests initializing the student as a model of the same size as the teacher and then pruning it during training.
%Knowledge distillation은 teacher의 knowledge를 받아 student의 성능을 증가시키는 방법으로 model compression technique의 일종이다. 대부분의 연구들은 text classification task를 위해 제안되었는데, 단순히 teacher와 student의 class distribution을 matching하는 연구부터 모델 사이의 hidden state나 attention matrix까지 matching하는 연구로 발전해왔다. 최근에는 큰 모델이 더욱 좋은 student를 만들어내지 않는다는 observation과 함께 teacher가 student의 capacity를 unaware한채 고정되어 knowledge를 전수한다는 점을 문제로 들며, student-friendly knowledge를 전수하는 방법들이 제안되고 있다. student-friendly knowledge를 추출하기 위해 distill 과정에서 teacher 역시 변형되게 되는데, 특정 objective를 통해 student의 영향을 받아 변형된다. 학습 초기 Teacher와 Student의 capacity 차이에 집중하여, Teacher를 점점 pruning하거나, Student를 teacher와 같은 크기의 model로 init하여 pruning해가는 방법도 제안되었다. 

\paragraph{KD for text generation}

For text generation, \citet{sanh2019distilbert} minimizes the KL divergence between the next token prediction distributions of the teacher and student at each time step. In addition, some research~\citep{calderon-etal-2023-systematic,gkd} focus on the sentences inputted to the teacher and student during the distillation process. For example, \citet{kim-rush-2016-sequence} uses sentences generated by the teacher as pseudo-targets instead of ground truth. Moreover, black-box KD methods~\citep{hsieh-etal-2023-distilling,ho-etal-2023-large} that use inference-only black-box LLMs as teachers and augment existing data before training are proposed. Recently, \citet{gkd,minillm} explored discrepancy metrics between model distributions and used sentences generated by the student as pseudo-targets to minimize exposure bias. However, there have been no attempts yet to distill student-friendly knowledge while the teacher is aware of the student's capacity. Although \citet{liang2023less} incorporates task-aware filters into both teacher and student to transfer knowledge, its scalability is limited due to the addition of filters at each layer for layer distillation. Crucially, it encourages knowledge to be task-specific, making it diverge from what we aim to explore in this paper. 
% For text generation, WordKD는 각 time step에서 teacher와 student의 next token prediction distribution 사이의 KL divergence를 최소화한다. Distill 과정에서 Teacher와 Student에 입력되는 문장에 집중하는 연구도 있는데 SeqKD는 GT가 아닌 teacher가 생성한 문장을 PT로 사용하며, inference만 가능한 black-box LLM을 teacher로 사용하여 기존의 data를 보강한 뒤 학습에 사용하는 black-box KD 방법들도 있다. 최근 GKD와 MiniLLM은 model distribution 사이의 discrepancy metric를 explore하였으며, input data 역시 exposure bias를 최소화하기 위해 Student가 생성한 문장을 PT로 사용한다. 하지만, 아직 teacher가 student의 capacity를 aware한채 student-friendly knowledge를 distill하는 시도는 없다. 비록 TED가 task-aware filter를 teacher와 student에 모두에 추가하여 student에게 knowledge를 전수하지만, 각 layer에서 filter를 추가하여 distill하므로 scalable하지 않다는 단점과 함께 결정적으로 knowledge가 task-specific하도록 encourage한다는 점에서 우리 연구와 비교하기 힘들다.

\paragraph{Prompt tuning}

After \citet{brown2020language} demonstrates that pre-trained language models can perform specific tasks by prepending text prompts to input, many studies have tried to either manually craft~\citep{schick-schutze-2021-exploiting} or automatically discover~\citep{shin-etal-2020-autoprompt,jiang-etal-2020-know,gao-etal-2021-making} such hard prompts, which are discrete tokens. Subsequently, research~\citep{hambardzumyan-etal-2021-warp,zhong-etal-2021-factual} emerged to advance prompts into the form of soft prompts composed of embeddings, making prompt updates via back-propagation easier and resulting in better performance compared to hard prompts. Presently, prompt tuning~\citep{lester-etal-2021-power} has become a prominent parameter-efficient fine-tuning technique. Although \citet{ijcai2022p596} uses hard prompts to generate input data for knowledge extraction, we are pioneering the use of prompts for parameter-efficient fine-tuning in KD research.
%Additionally, as the size of the model increases, the difference between prompt tuning and full-parameter fine-tuning decreases~\citep{lester-etal-2021-power}, suggesting that our method could yield even greater benefits when applied to large language models.
% Text prompt를 input에 prepend하면 Pre-trained language model이 특정 task를 수행할 수 있다는 점이 밝혀지자, 이러한 prompt를 사람이 읽을 수 있는 단어 형태로 직접 만들거나 automatically 찾는 연구들이 있었다. 그 후, prompt를 사람이 해석할 수 없는 embedding으로 이루어진 soft prompt의 형태로 발전시킨 연구들이 등장했는데, gradient에 의한 update가 용이해져서 hard prompt일때보다 더욱 좋은 prompt를 찾을 수 있게 되었다. 현재 prompt tuning은 prominent한 parameter-efficient fine-tuning 기법으로 자리매김하였다. knowledge distillation 연구에서 knowledge 추출에 사용될input data를 생성하기 위해 hard prompt를 사용하는 연구는 있지만, model을 fine-tuning하기 위해 prompt tuning을 활용한 것은 이것이 첫 시도이다. 모델의 크기가 커질수록 prompt tuning과 full-parameter fine-tuning의 차이가 줄어든다는 observation도 있기 때문에, 우리 방법이 LLM에 적용한다면 더욱 큰 효과를 기대할 수 있다.

\begin{figure*}[!t]
\centering
\includegraphics[width=\linewidth]{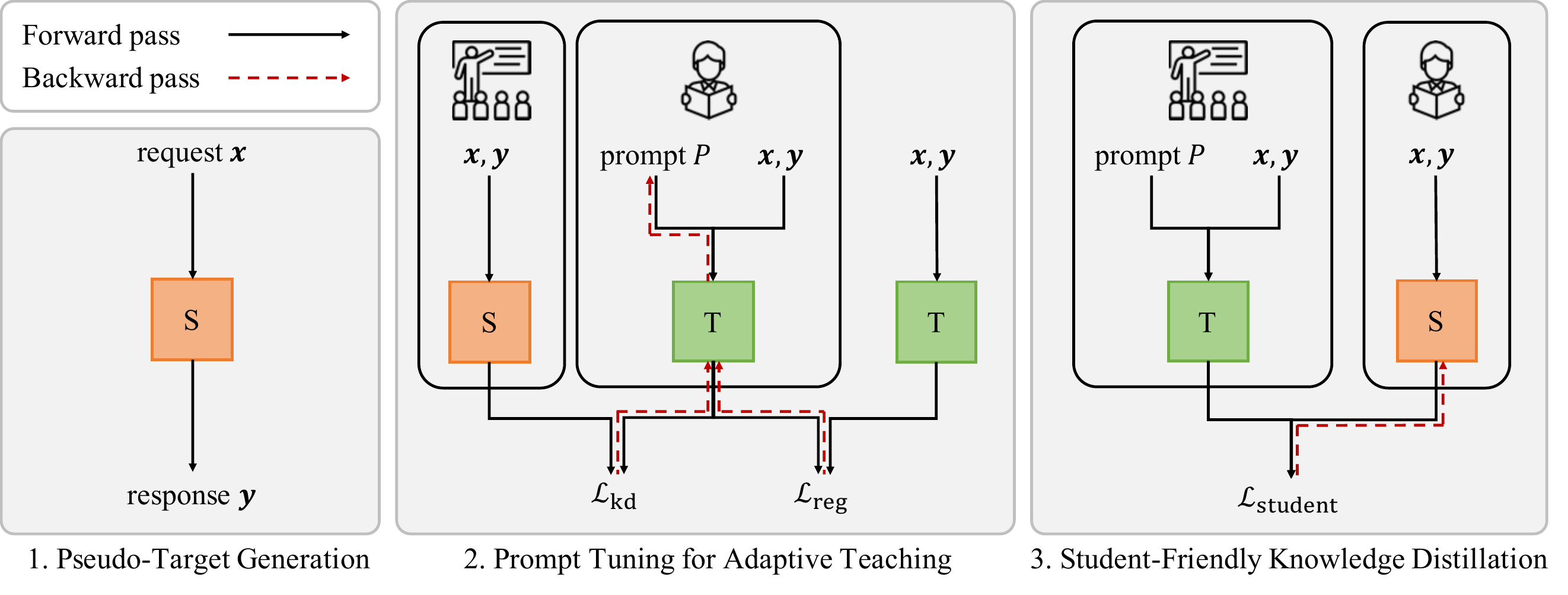}
\caption{Training procedure of PromptKD. To mitigate exposure bias, responses are generated by the student to be used as pseudo-targets. Then, for adaptive teaching, the prompt input to the teacher is trained based on guidance from the student. During this process, regularization loss is also employed to address instability stemming from the prompt. Lastly, teacher distills student-friendly knowledge to the student using the trained prompt.}
%Different from the previous KD method, our method updates prompt with the student model's feedback to generate student-friendly knowledge by teacher model. With this additional step, PromptKD achieves a state-of-the-art performance on the instruction following dataset.
\label{fig:main_figure}
\end{figure*}

\section{PromptKD}

PromptKD is devised in the instruction-following~\citep{ouyang2022training} setting for application to generative language models. We formulate instruction-following as a conditional text generation task, where the request $\bm{x}=\{x_1,x_2,\dots,x_n\}$ sampled from the data distribution $p_x$ comprises instruction and input to describe the task. Then, given the request $\bm{x}$ as a condition, the model generates a response $\bm{y}=\{y_1,y_2,\dots,y_T\}$. For prompt tuning, soft prompts $P=\{\bm{p}_1,\bm{p}_2,\dots,\bm{p}_m\}$, where $\bm{p}_i$ is an embedding vector of the same dimension as the token embedding, are initialized with text and prepended to the input request $\bm{x}$. Formally, given the request $\bm{x}$, the teacher model distribution conditioned on the prompt $P$ is denoted as $p(\bm{y}|P,\bm{x})$ (here we suppress the teacher's model parameter since it is fixed), and the student's model distribution parameterized by $\theta$ is denoted as $q_\theta(\bm{y}|\bm{x})$, where only the student model parameters $\theta$ and the prompt $P$ are trainable. The training process consists of 3 steps per iteration, as shown in Figure~\ref{fig:main_figure}. First, generating input data used for knowledge distillation (\emph{pseudo-target generation}). Then, updating the prompt based on guidance from the student and teacher models to facilitate adaptive teaching (\emph{prompt tuning for adaptive teaching}). Finally, distilling student-friendly knowledge to the student using the updated prompt (\emph{student-friendly knowledge distillation}).
% PromptKD는 LLM과 같은 generative language model에 적용될것을 고려하여 instruction-following setting에서 고안되었다. conditional text generation task로 formulate하면, data distribution px에서 sampling되는 request x는 instruction과 input으로 구성되어 task를 설명하는 역할을 한다. Then, request x를 조건으로 model은 response y={y1, …, yT}를 생성한다. 또한 Teacher T를 prompt tuning하기 위해서 Text로 initialize된 soft prompt P={p1, …, pl}, where p_i은 input dimension과 같은 차원을 가진 embedding vector,를 input인 request x에 prepend하여 teacher에 입력한다. Formally, request x를 받았을때, teacher model distribution은 p(y|p,x)로, theta로 parametrized 된 student의 model distribution은 q_theta(y|x)로 denote한다. student model parameter theta와 prompt P만 learnable하다. 학습 과정에서의 매 iteration은 Figure 1에 보이는 것처럼 총 3단계로 구성된다. 먼저, knoweldge 추출에 사용되는 input data를 생성한다 *(pseudo-target generation)*. Then, student와 teacher model의 guidance를 통해 prompt를 update한다 (*prompt update*). 마지막으로, update된 prompt로 student-friendly knoweldge를 추출하여 student에게 distil한다 (*student-friendly knowledge distillation*).

\subsection{Pseudo-Target Generation}

PromptKD uses the response $\bm{y}$ generated by the student for the prompt tuning and knowledge distillation processes, treating it as the pseudo-target. This approach addresses exposure bias, which arises due to the discrepancy between the sentences used during training and those generated during inference, leading to degraded performance in free-run generation~\citep{zhang-etal-2019-bridging}. Based on the insight~\citep{gkd} that incorporating sentences that the model can generate during free-run generation into the training process can mitigate exposure bias, we devise the approach accordingly. It is worth noting that for the sake of method simplicity, back-propagation during this sampling process is not conducted. 
% While \citet{gkd,minillm} propose mixing the student's results with the teacher or ground truth as the pseudo-target, \citet{minillm} pursues this due to reward hacking concerns and \citet{gkd} showes the best results without mixing. Hence, we solely relied on the student for pseudo-target generation.
% 우리는 request x에 대해 Student가 생성한 response y를 prompt update와 knowledge distillation 과정 모두에서 pseudo-target으로 사용하여, teacher-forcing 방식으로 입력하면서 output distribution에 대해 loss를 측정한다. **[수식]**  이는 exposure bias 때문인데, exposure bias는 학습에 사용된 문장과 model이 inference때 생성할 수 있는 문장 사이의 discrepancy로 발생하며 클 경우 model의 free-run generation 성능을 떨어뜨린다. model이 free-run generation때 생성할 수 있는 문장을 학습에 사용하면 이러한 exposure bias를 줄일 수 있다는 observation이 있어 student가 생성한 문장을 사용했다. 이때, method의 simplicity를 위해 sampling 과정에서의 back-propagate은 일어나지 않는다. 두 연구들 모두 Pseudo-Target으로 Student의 결과를 Teacher 또는 Ground Truth와 섞는 방법을 제안하였으나, MiniLLM은 reward hacking 때문에 진행한 것이며, GKD는 섞지 않았을때 가장 좋은 결과를 보여주었기 때문에, 우리는 Student에만 의존하여 Pseudo-Target을 생성했다. 

\subsection{Prompt Tuning for Adaptive Teaching}

Initially, we concatenate the request $\bm{x}$ and response $\bm{y}$, including the prompt $P$ for the teacher, and input them into both models. Prompt $P$ is updated to minimize the KD loss $\mathcal{L}_{\text{kd}}$, which computes the distribution discrepancy of the response part. This encourages the prompt to enable the teacher to generate sentences at a similar level to the student when it is prepended to the teacher's input. Drawing inspiration from the concept of adaptive teaching in education, we design this objective with the aim of enabling students to receive knowledge from the teacher at a level they can comprehend.
%We design this objective to facilitate the extraction of knowledge that is more easily understandable for the student from a fixed teacher.
% request x와 생성된 Pseudo-Target y을 포함하여 각 teacher와 student에 입력한 뒤 teacher-forcing 방식으로 next token prediction을 수행한다. teacher에는 prompt p가 prepend된 채 입력된다. response에 해당하는 부분의 distribution discrepancy를 계산한 KD loss를 최소화되도록 prompt를 학습함으로써, prompt가 teacher의 입력에 prepend되었을 때 teacher가 student와 유사한 수준의 문장을 생성할 수 있도록 encourage한다. 이 objective는 기존의 고정된 Teacher보다, student가 더욱 이해하기 쉬운 student-friendly knowledge를 추출할 수 있도록 고안되었다.

However, during the early stages of training, the influence of the prompt may cause significant deviations or inaccuracies in the teacher model distribution, leading to unstable learning~\citep{hou-etal-2022-metaprompting}. To address this issue, we initialize the prompt with text embedding and devise an additional regularization loss $\mathcal{L}_{\text{reg}}$ to ensure that the teacher model distribution remains similar whether the prompt is used or not. The regularization loss $\mathcal{L}_{\text{reg}}$ is computed in a similar manner to the KD loss $\mathcal{L}_{\text{kd}}$, but with the difference that it is measured based on the teacher model distribution when the prompt is excluded from the input given to the teacher. This approach allows for the continued use of the fixed teacher model, making it memory-efficient. However, since the fixed teacher is unaware of the student's capacity, $\mathcal{L}_{\text{reg}}$ deviates from our ultimate goal. Therefore, we introduce a coefficient that starts at 1 for $\mathcal{L}_{\text{reg}}$ and linearly decreases to 0 during training, focusing solely on stabilizing the early stages of learning.
% 그러나, 학습 초기에 prompt의 영향으로 teacher model distribution이 기존과 크게 달라지거나 부정확할 수 있으며, 이는 불안정한 학습을 초래한다. To tackle this, 우리는 text로 prompt를 initialize하였고, prompt를 사용할때와 그렇지 않을 때 teacher model distribution이 서로 비슷해지도록 하는 loss를 추가로 고안하였다. teacher에 입력된 data에서 prompt만을 빼고 다시 입력하여 model distribution을 구한 뒤 똑같이 response part에 대해 distribution discrepancy를 측정하는 loss로, teacher model은 그대로 사용할 수 있어 memory-efficient하다. 하지만, 고정된 teacher는 student의 capacity를 unaware하기 때문에, 우리의 최종 목적과는 거리가 있다. 따라서, 우리는 이러한 loss에 1로 시작하여 최종적으로 0까지 학습 과정에서 linear하게 줄어드는 계수를 붙여서 학습 초반의 안정성만을 키워주었다. 

Regarding the two objectives, we opt for minimizing the reverse KL divergence instead of the forward KL divergence to measure the discrepancy, as it exhibits mode-seeking behavior~\citep{nowozin2016f} and benefits generation tasks. Hence, summarizing the two objectives, the final loss $\mathcal{L}_{\text{prompt}}$, which updates only the prompt, is determined by their summation, as follows:
\begin{align}
    %\mathcal{L}_{\text{KD}}=&\mathop{\mathbb{E}}_{\bm{y}\sim q_\theta(\cdot|\bm{x})}\left[   \frac{1}{T}\sum_{t=1}^T D_{KL}\left( p(\cdot|\bm{y}_{<t},\bm{x}) \middle\| q_\theta(\cdot|\bm{y}_{<t},\bm{x}) \right)  \right],\\
    \mathcal{L}_{\text{kd}}=&D_{KL}\big( p(\bm{y}|P,\bm{x}) \parallel q_\theta(\bm{y}|\bm{x}) \big), \\[5pt]
    \mathcal{L}_{\text{reg}}=&D_{KL}\big( p(\bm{y}|P,\bm{x}) \parallel p(\bm{y}|\bm{x}) \big), \\[5pt]
    \mathcal{L}_{\text{prompt}}=&\mathcal{L}_{\text{kd}}+\left(\frac{K-k}{K}\right)\mathcal{L}_{\text{reg}},
    \label{eqn:prompt}
\end{align}
where $K$ represents the total training steps, and $k$ denotes the current step.
% 두 가지 objective와 관련하여, forward KL divergence가 아닌 reverse KL divergence로 discrepancy를 측정하여 최소화했을 때 mode-seeking behavior가 보여지며 generation에 유리해져서 reverse KL divergence를 선택했다. 따라서, 두 objective를 정리하면 아래와 같으며, 이 둘을 더한 최종 L_prompt에 의해 prompt만 update된다. **[수식]** , where K는 total training step, k는 현재 step.

\begin{algorithm}[!t]
\caption{PromptKD}
\label{alg:promptkd}
\begin{algorithmic}
\Require teacher $T$, student's output distribution $q_\theta$, data distribution $p_x$, prompt $P$, training step $K$, learning rate $\eta$
\For{each step $k=1,...,K$}
    \State Sample a request $\bm{x}$ from $p_x$
    \State Sample a response $\bm{y}$ from $q_\theta(\cdot|\bm{x})$
    \State Update $P \leftarrow P-\eta \nabla \mathcal{L}_{\text{prompt}}$  \quad $\triangleright$ Eq.~\eqref{eqn:prompt}
    \State Update $\theta \leftarrow \theta-\eta \nabla \mathcal{L}_{\text{student}}$  \quad $\triangleright$ Eq.~\eqref{eqn:student}
\EndFor
\State \Return $q_\theta$
\end{algorithmic}
\end{algorithm}

\begin{table*}[!t]
\centering
\begin{tabular}{lrlccccc}
\hline
\multirow{2}{*}{Model} & \multirow{2}{*}{\#Params} & \multirow{2}{*}{Method} & \multicolumn{5}{c}{Instruction-following datasets} \\ 
\cline{4-8} &  &  & Dolly & SelfInst & Vicuna & S-NI & UnNI \\ 
\hline
\hline
\multirow{19}{*}{GPT-2} 
& 1.5B  & Teacher & 27.3 & 14.5 & 16.2 & 27.1 & 31.6 \\ 
\cline{2-8}
& \multirow{6}{*}{120M} 
& SFT & 22.9 & 10.2 & 14.5 & 16.3 & 18.5 \\
& & KD & 22.6 & 11.0 & 15.1 & 18.0 & 20.1 \\
& & SeqKD & 23.3 & 10.3 & 14.7 & 16.6 & 19.2 \\
& & GKD & 24.8 & 11.1 & \textbf{17.7}$^\dagger$ & 20.7 & 23.2 \\
& & MiniLLM & 24.2 & 12.7 & 16.9$^\dagger$ & 25.1 & 26.2 \\
& & PromptKD & \textbf{25.6} & \textbf{13.1} & 16.8$^\dagger$ & \textbf{26.8} & \textbf{28.9} \\
\cline{2-8}
& \multirow{6}{*}{340M} 
& SFT & 25.1 & 12.9 & 15.9 & 23.7 & 27.4 \\
& & KD & 25.1 & 13.0 & 15.6 & 24.5 & 27.7 \\
& & SeqKD & 25.3 & 12.7 & 16.0 & 23.8 & 27.5 \\
& & GKD & 26.9 & 14.8$^\dagger$ & 17.8$^\dagger$ & 26.6 & 30.9 \\
& & MiniLLM & 26.3 & 14.8$^\dagger$ & \textbf{17.9}$^\dagger$ & 26.4 & 31.2 \\
& & PromptKD & \textbf{27.3}$^\dagger$ & \textbf{15.0}$^\dagger$ & 17.6$^\dagger$ & \textbf{27.1}$^\dagger$ & \textbf{32.6}$^\dagger$ \\
\cline{2-8}
& \multirow{6}{*}{760M} 
& SFT & 24.9 & 13.4 & 15.8 & 24.0 & 27.6 \\
& & KD & 25.7 & 13.7 & 15.9 & 24.0 & 27.7 \\
& & SeqKD & 25.2 & 13.3 & 15.8 & 24.0 & 27.4 \\
& & GKD & \textbf{26.9} & 14.1 & 17.1$^\dagger$ & 25.4 & 29.6 \\
& & MiniLLM & 26.2 & 15.8$^\dagger$ & 16.9$^\dagger$ & 28.5$^\dagger$ & 33.5$^\dagger$ \\
& & PromptKD & \textbf{26.9} & \textbf{16.4}$^\dagger$ & \textbf{17.8}$^\dagger$ & \textbf{29.5}$^\dagger$ & \textbf{34.8}$^\dagger$ \\
\hline
\multirow{7}{*}{OPT} 
& 13B & Teacher & 29.3 & 17.7 & 17.3 & 30.7 & 33.8 \\ 
\cline{2-8}
& \multirow{2}{*}{1.3B} 
& MiniLLM & 26.8 & 15.2 & 18.1$^\dagger$ & 28.6 & 30.9 \\
& & PromptKD & \textbf{28.0} & \textbf{15.5} & \textbf{18.5}$^\dagger$ & \textbf{29.6} & \textbf{33.5} \\
\cline{2-8}
& \multirow{2}{*}{2.7B} 
& MiniLLM & 27.2 & 16.2 & 18.6$^\dagger$ & 29.8 & 33.1 \\
& & PromptKD & \textbf{28.7} & \textbf{17.8}$^\dagger$ & \textbf{18.9}$^\dagger$ & \textbf{31.4}$^\dagger$ & \textbf{34.8}$^\dagger$ \\
\cline{2-8}
& \multirow{2}{*}{6.7B}
& MiniLLM & 28.6 & 18.0$^\dagger$ & 19.1$^\dagger$ & 32.5$^\dagger$ & 34.5$^\dagger$ \\
& & PromptKD & \textbf{29.9}$^\dagger$ & \textbf{19.0}$^\dagger$ & \textbf{19.8}$^\dagger$ & \textbf{33.8}$^\dagger$ & \textbf{35.2}$^\dagger$ \\
\hline
\multirow{3}{*}{Llama} 
& 13B & Teacher & 30.2 & 23.1 & 19.0 & 35.7 & 36.9 \\ 
\cline{2-8}
& \multirow{2}{*}{7B} 
& MiniLLM & 29.0 & 21.3 & 20.6$^\dagger$ & \textbf{36.7}$^\dagger$ & 38.1$^\dagger$ \\
& & PromptKD & \textbf{30.0} & \textbf{23.4}$^\dagger$ & \textbf{21.1}$^\dagger$ & 36.6$^\dagger$ & \textbf{38.9}$^\dagger$ \\
\hline
\end{tabular}
\caption{Evaluation results on 5 instruction-following datasets. Each ROUGE-L score is averaged over 5 random seeds. The best score for each model size is highlighted in \textbf{boldface}. $^\dagger$Results surpass those of the teacher.}
\label{tab:main_result}
\end{table*}

\subsection{Student-Friendly Knowledge Distillation}

The updated prompt is utilized as a trigger to extract student-friendly knowledge from the teacher and distill it to the student. The student loss $\mathcal{L}_{\text{student}}$ minimizes the distribution discrepancy between teacher and student through reverse KL divergence, as follows:
\begin{align}
    \mathcal{L}_{\text{student}}=&D_{KL}\big( q_\theta(\bm{y}|\bm{x}) \parallel p(\bm{y}|P,\bm{x}) \big).
    \label{eqn:student}
\end{align}
%The reason for selecting the reverse KL among various KL-based divergences is that it tends to have fewer modes due to the relatively smaller capacity, thereby suppressing the occurrence of probabilities on tokens predicted with very low probability by the teacher and encouraging mode-seeking behavior (minillm citation list). 
For a clear understanding, we summarize the PromptKD algorithm in Algorithm~\ref{alg:promptkd}.
% Update된 prompt를 trigger처럼 활용하여 teacher로부터 student-friendly knowledge를 추출하여 student에 전수한다. Student를 학습시키는데 사용되는 student loss L_stu 역시 reverse KL divergence를 통해 distribution discrepancy를 최소화한다. 이러한 objective L_kd를 수식으로 표현하면 아래와 같으며, student model parameter인 theta가 학습된다. **[수식]** 여기서 다양한 KL-based divergences중 reverse KL을 선택한 이유는, 상대적으로 capacity가 작아 더 적은 수의 mode를 가지기 때문에, teacher가 아주 작은 확률로 예측한 토큰에 확률이 발생하는 것을 억제하고 mode-seeking behavior를 장려하기 위해서이다 (minillm cite 목록). Clear understanding을 위해 PromptKD를 algorithm 1에 요약한다.

\section{Experiments}

\subsection{Experimental Setup}

Following \citet{minillm}, we evaluate PromptKD using 5 instruction-following datasets.
% Following previous works(MiniLLM), 우리는 PromptKD를 5가지 instruction-following dataset들로 evaluate했다. 

\paragraph{Settings}

We split the Dolly~\citep{databrick2023dollyv2}, consisting of 15,000 human-written instruction-response pairs, into 14,000 for training and 500 for validation and testing. For evaluation, we employ not only the Dolly but also 4 additional datasets: SelfInst~\citep{wang-etal-2023-self-instruct}, consisting of user-oriented instruction-following sets; Vicuna~\citep{chiang2023vicuna}, comprising 80 questions used in the Vicuna evaluation; S-NI, the test set of \textsc{Super-NaturalInstructions}~\citep{wang-etal-2022-super}; and UnNI, the core dataset of \textsc{UnnaturalInstructions}~\citep{honovich-etal-2023-unnatural}. Similar to \citet{minillm}, data samples with ground truth response lengths of 11 or more are utilized for S-NI and UnNI. We generate 5 responses for each request in each dataset using different random seeds and evaluate them to report the average scores for reliability. We choose the ROUGE-L score~\cite{lin-2004-rouge} as the metric for evaluation, as it aligns well with human preferences~\citep{wang-etal-2022-super} in instruction-following evaluations. The best checkpoint based on the ROUGE-L score on the validation set is used for evaluation. We also measure the GPT-4 feedback scores~\citep{zheng2024judging}, which are separately summarized in Appendix~\ref{sec:gpt4_score}. 
% 15K개의 human-written instruction-response pair로 구성된 dolly dataset을 split하여, 14K개를 학습 데이터 사용하면서 500개의 sample을 validation과 test 용으로 구성하였다. evaluation을 위해서 dolly dataset 뿐만 아니라 4개의 dataset을 더 사용했는데, user-oriented instruction-following set인 SelfInst와 Vicuda evaluation에 사용된 80개의 질문으로 구성된 Vicuna, SUPER-NATURALINSTRUCTIONS의 test set인 S-NI, UNNATURALINSTRUCTION의 core set인 UnNI 이다. MiniLLM과 같이 S-NI와 UnNI의 경우 ground truth response 길이가 11 이상인 data sample을 사용했다. 각 dataset의 request에 대해 서로 다른 random seed로 총 5번 response를 생성한 뒤, 이를 평가하여 평균낸 score를 report한다. metric의 경우 ROUGE-L score가 instruction-following evaluation에 있어 human preference와 잘 align됨으로써 적합하다는 observation이 있어 이를 사용했다. evaluation을 위한 model checkpoint 역시 이 ROUGE-L score validation set에 대해 가장 좋았던 checkpoint를 사용했다. 

% \NEW{GPT-4 feedback scores도 측정해 보았는데, 이는 appendix A에 정리했다.}

\paragraph{Models}

To evaluate the instruction-following performance of PromptKD across various models, we utilize pre-trained GPT-2~\citep{radford2019language}, OPT~\citep{zhang2022opt}, and Llama~\citep{touvron2023llama1} model families. For the GPT-2 model family, GPT-2 XL (1.5B params) is employed for the teacher model, and GPT-2 Base (120M params), GPT-2 Medium (340M params), GPT-2 Large (760M params) are used for the student model. For the OPT and Llama model families, we use OPT-13B and Llama-13B as the teacher models, and OPT-1.3B, OPT-2.7B, OPT-6.7B, and Llama-7B as the student models, respectively.
%GPT-2 XL (1.5B params) is employed for the teacher model, and GPT-2 Base (120M params), GPT-2 Medium (340M params), GPT-2 Large (760M params) are used for the student model. 
Before knowledge distillation, the teacher model undergoes supervised fine-tuning on the Dolly training set. Similarly, the student model is also fine-tuned on the same training data for only three epochs, following the previous works~\citep{gkd,minillm}.
% PropmtKD의 instruction-following performance를 평가하기 위해 pre-trained된 GPT-2 model family를 사용한다. Teacher model의 경우 GPT-2-1.5B를 사용했고, Student model은 GPT-2 (120M, 340M, 760M) 모델을 사용했다. Knowledge distillation을 수행하기 전에, teacher의 경우 Dolly training set으로 supervised fine-tuning했으며, student도 previous works와 마찬가지로 동일한 학습 데이터에 대해 3 epoch만 supervised fine-tuning했다.
% \NEW{PropmtKD의 instruction-following performance를 다양한 모델에서 평가하기 위해 pre-trained된 GPT-2, OPT, Llama model family들을 사용한다. GPT-2 model family의 경우, Teacher model은 GPT-2-1.5B를 사용했고, Student model은 GPT-2 (120M, 340M, 760M) 모델을 사용했다. OPT model family는 teacher model로 OPT-13B, student model로 OPT-1.3B, OPT-2.7B, OPT-6.7B를 사용한다. Llama의 경우 teacher model은 Llama-13B, student model로는 Llama-7B를 사용한다. Knowledge distillation을 수행하기 전에, teacher의 경우 Dolly training set으로 supervised fine-tuning했으며, student도 previous works와 마찬가지로 동일한 학습 데이터에 대해 3 epoch만 supervised fine-tuning했다.}

\paragraph{Baselines}

PromptKD is compared with various approaches ranging from supervised fine-tuning (SFT), which does not involve knowledge distillation, to commonly used methods in generation tasks such as Supervised KD (KD; \citealp{sanh2019distilbert}), SeqKD~\citep{kim-rush-2016-sequence}, and more recent proposals like MiniLLM~\citep{minillm} and GKD~\citep{gkd}. KD and SeqKD both aim to minimize the discrepancy between the model distributions of teacher and student at each token step. The difference lies in whether the input sentence is ground truth or pseudo-target generated by the teacher. MiniLLM replaces forward KL divergence with reverse KL divergence and updates the student model using policy gradient. On the other hand, GKD focuses on distribution discrepancy metrics and pseudo-targets to propose a general method. In this paper, GKD computes reverse KL divergence and utilizes sentences generated by the student as pseudo-targets, and this choice is based on the reported performance in their paper. Additionally, it is worth noting that the students for MiniLLM, GKD, and PromptKD all commence from the same supervised fine-tuned checkpoint, while other methods start from pre-trained models. Due to resource limitations, experiments on the OPT and Llama models are conducted only in comparison with MiniLLM, which demonstrated outstanding performance among all baselines in the GPT-2 results. For training details, please see the Appendix~\ref{sec:training_detail}.
% PromptKD는 KD를 진행하지 않는 supervised fine-tuning부터 generation task를 위한 KD에서 많이 사용되는 SeqKD, WordKD, 그리고 최근에 제안된 MiniLLM과 GKD와 비교된다. WordKD와 SeqKD 모두 각 token step에서 teacher와 student의 model distribution 사이의 discrepancy를 최소화하는 방법이며, 입력되는 문장이 ground truth인지 Teacher가 생성한 Pseudo-Target인지의 차이가 있다. MiniLLM은 forward KL divergence를 reverse KL divergence로 대체하여 policy gradient와 함께 student를 update하는 방법이며, GKD 역시 distribution discrepancy metric과 pseudo-target에 집중하여 일반적인 방법을 제안했다. In this paper, GKD의 경우 report한 성능을 기반으로 metric은 reverse KL divergence, Pseudo-target은 student가 생성한 문장을 사용했으며, MiniLLM과 GKD, PromptKD는 동일한 supervised fine-tuned checkpoint에서 시작한다. 

%\NEW{본 논문에서는 hardware resource의 한계로, OPT와 Llama model의 실험은 모든 GPT-2 student model에서 baseline 중 가장 높은 평균 성능을 일관적으로 보여준 MiniLLM과의 비교만 수행한다.}

\subsection{Experimental Results}

We report the instruction-following performance of PromptKD and baselines on 5 datasets in Table~\ref{tab:main_result}.
% 우리는 5가지 dataset에 대한 instruction-following performance를 Table에 report했다. 

Firstly, PromptKD achieves state-of-the-art performance overall in the instruction-following setting, outperforming other KD baselines. Additionally, it also outperforms on 4 datasets not used in training, demonstrating PromptKD's superb generalization ability. These results robustly demonstrate the superiority of PromptKD, as they consistently appear across all model families and model sizes. It's worth noting that despite MiniLLM incorporating language modeling loss through the corpus used for pre-training, PromptKD exhibits better performance. 
% 먼저, PromptKD는 instruction-following setting에서 전체적으로 state-of-the-art performance를 달성했으며 다른 KD baseline들에 비해 outperform했다. 또한 학습에 사용되지 않은 4개의 dataset에서도 outperform했는데, 이는 PropmtKD의 뛰어난 generalization ablity를 증명한다. Please note that MiniLLM은 pre-training에 사용되는 corpus를 통해 language modeling loss를 함께 사용했음에도 PromptKD가 더욱 좋은 성능을 보였다. 
% \NEW{이 결과는 모든 model family와 student size에서 동일하게 나타났기 때문에 더욱 PromptKD의 우수성을 강력하게 입증하는 결과이다.}

Furthermore, only PromptKD shows superior performance to the teacher across all datasets. This demonstrates that modifying the teacher to extract student-friendly knowledge for distillation works not only for classification tasks but also for generation tasks. Moreover, the better performance of PromptKD, MiniLLM, and GKD, which utilize responses generated by the student as pseudo-targets, compared to other baselines, can be interpreted as exposure bias mitigation contributing to the performance improvement. 
% 또한, PromptKD는 모든 dataset에서 teacher보다 좋은 성능을 보이기도 했다. 이를 통해 teacher를 변형시켜 student-friendly knowledge를 추출하여 distill하는 것이 classification task 뿐만 아니라 generation task에서도 working하는 것을 demonstrate한다. 게다가 student가 생성한 response를 pseudo-target으로 사용하는 PromptKD, MINILLM, GKD가 other baseline들보다 좋은 성능을 보여준 것은 이러한 방식이 exposure bias 해소에 도움이 되었기 때문으로 추측할 수 있겠다.

% Lastly, as the model size increases, PromptKD outperforms in more datasets. This can be attributed to the fact that prompt tuning exhibits a similar effect to full-parameter fine-tuning as the model size grows~\citep{lester-etal-2021-power}. Due to the scalability and efficiency of prompt tuning, PromptKD can be expected to yield outstanding results even when applied to larger models.
% % 마지막으로, model의 크기가 커질수록 더욱 많은 dataset에서 PromptKD가 outperform했는데 이는 model이 커질수록 prompt tuning이 full-parameter fine-tuning과 유사한 효과를 보였기 때문으로 추측할 수 있다. prompt tuning의 scalable하며 efficient함 덕분에, PromptKD는 더욱 큰 모델에 적용했을 때도 뛰어난 효과를 기대할 수 있다.

PromptKD and the baselines' qualitative results are summarized in the Appendix~\ref{sec:qual_res}, where it is shown that PromptKD generates responses most similar to the ground truth.

\begin{figure*}[!t]
\centering
\subfigure[exposure bias against generation steps]{
    \includegraphics[width=0.48\textwidth]{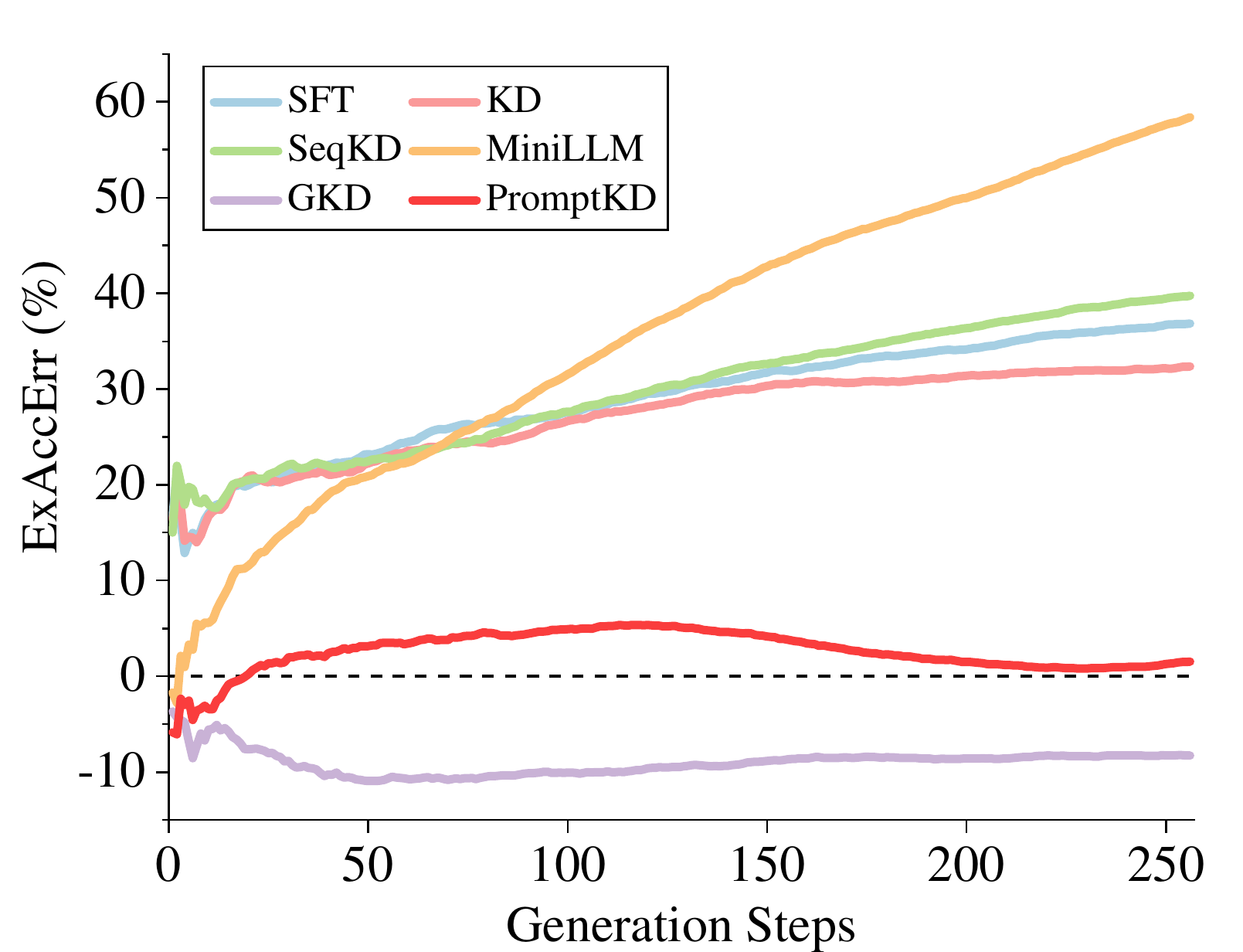}
    \label{fig:eb_length}
}
\subfigure[exposure bias against training progress]{
    \includegraphics[width=0.48\textwidth]{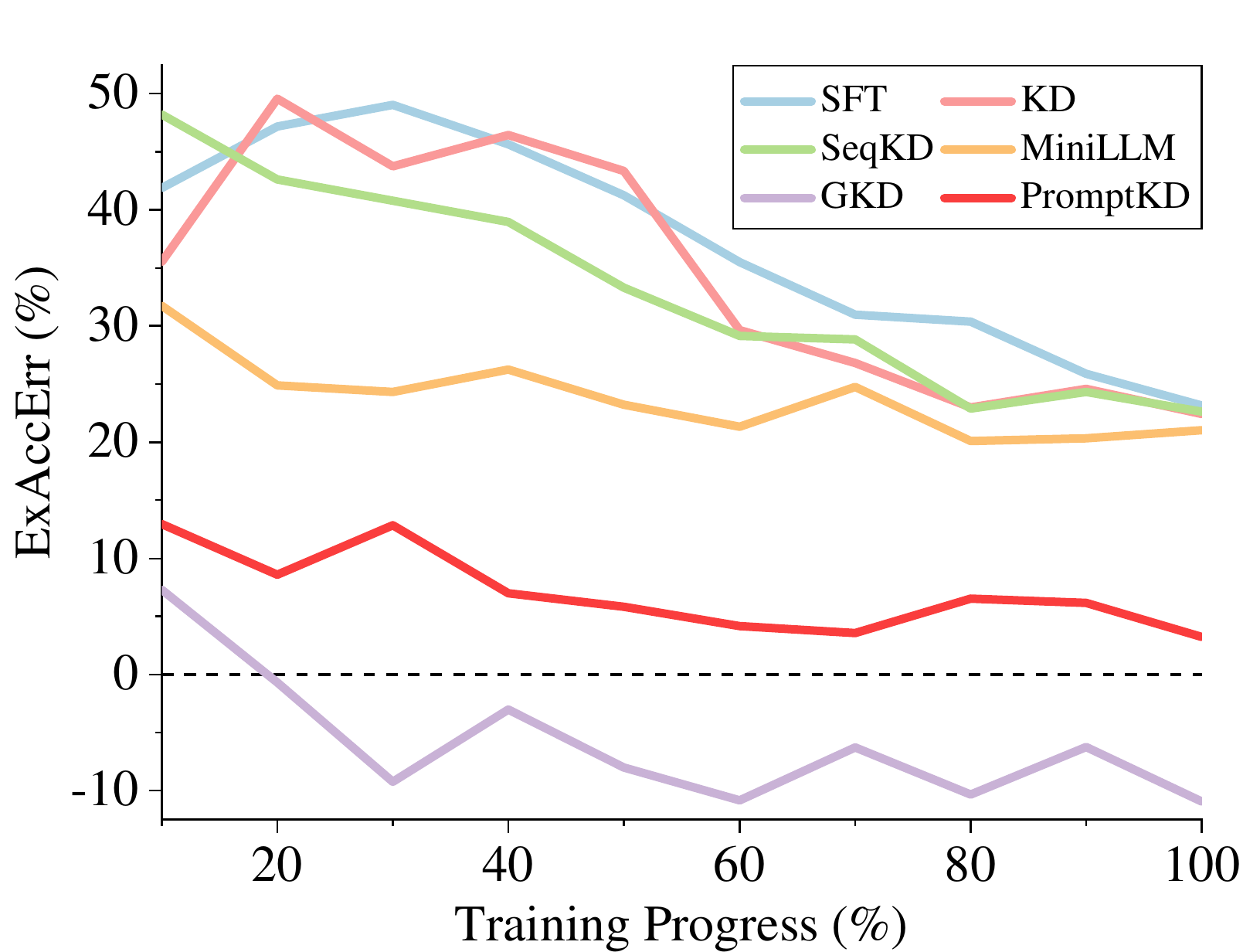}
    \label{fig:eb_training_step}
}
\caption{The measurement of exposure bias. Excess accumulated error (ExAccErr) is measured with respect to generation steps and training progress, where values closer to 0 indicate alleviation of exposure bias.}
%\textbf{Overview of our exposure bias} (a) Excess accumulated error (ExAccErr) plotted against generation length for the best-performing models. (b) ExAccErr until the model reaches its best performance. Training steps are evenly divided into 10 intervals up to the best-performing iteration.}
\label{fig:eb_fig}
\vspace{-0.1in}
\end{figure*}

\subsection{Analysis}

\paragraph{Exposure bias}

In this section, we investigate exposure bias to understand why PromptKD performs well. Exposure bias refers to the mismatch in distribution between the sentences seen during training and those generated during inference. If exposure bias is significant, the tokens generated during inference may diverge from those seen during training, leading to accumulated errors in the generation process. Following \citet{arora-etal-2022-exposure}, exposure bias up to $l$ generation steps can be quantified as follows: 
\begin{align}
    \text{ExAccErr}(l)=\frac{R(l)-E(l)}{E(l)} \times 100\%, \\
    R(l)=\sum_{t=1}^l\mathop{\mathbb{E}}\limits_{\substack{\bm{y}_{<t}\sim q_\theta(\cdot|\bm{x})\\y_t\sim p(\cdot|\bm{y}_{<t},\bm{x})}} \log{\frac{p(y_t|\bm{y}_{<t},\bm{x})}{q_\theta(y_t|\bm{y}_{<t},\bm{x})}}, \\   
    E(l)=\sum_{t=1}^l\mathop{\mathbb{E}}\limits_{\substack{\bm{y}_{<t}\sim p(\cdot|\bm{x})\\y_t\sim p(\cdot|\bm{y}_{<t},\bm{x})}} \log{\frac{p(y_t|\bm{y}_{<t},\bm{x})}{q_\theta(y_t|\bm{y}_{<t},\bm{x})}}.
\end{align}
$R(l)$ represents the average forward KL divergence up to $l$ time steps when the student-generated response is used as the pseudo-target, while $E(l)$ is similar to $R(l)$ but differs in that it uses the teacher-generated response as the pseudo-target. $R(l)$ can be interpreted as the distribution gap between the teacher and the student due to low-quality pseudo-targets generated by the student, while $E(l)$ serves as a lower-bound of distribution gap between the teacher and the student. Therefore, ExAccErr calculates the relative error caused solely by exposure bias. If exposure bias is alleviated, the student should exhibit a nearly identical distribution gap regardless of which model generated the response. Therefore, the ExAccErr value should approach 0.
% In this section, PropmtKD가 왜 잘 동작하는 것인지 분석하기 위해 PromptKD가 해소해줄 것으로 예상되는 exposure bias에 대해 investigate한다. exposure bias란 model이 학습때 본 문장과 inference때 생성하는 문장의 distribution 차이로, exposure bias가 크다면 inference 초기에 생성한 token들이 학습때 본것과 달라지게 되어 생성 과정에서 error가 누적되는 문제가 발생한다. Following MiniLLM, exposure bias를 아래와 같은 수식으로 수치화하였으며 R(l)은 Student가 생성한 response를 pseudo-target으로 사용했을 때의 forward KL divergence의 l time step까지의 평균, le(l)은 teacher가 생성한 response를 pseudo-target으로 사용했을 때의 forward KL divergence의 l time step까지의 평균이다. [수식]. teacher model을 oracle로 봤을 때, R(l)은 student가 생성한 low-quality pseudo-target으로 인한 error이고, le(l)는 oracle context가 주어졌을때 oracle model과 student 사이의 오차로 lower-bound와 같은 역할을 한다고 해석할 수 있다. 만약 exposure bias가 해소되었다면, student 입장에서는 어떤 모델이 생성한 response든지 상관없이 oracle model과 동일한 오차를 가져야 하므로, ExAccErr 값이 0에 가까워야 한다.  

We depict the ExAccErr at each generation step and the variation of ExAccErr up to 50 generation steps during the model training in Figure~\ref{fig:eb_fig}. In this experiment, a fixed pre-trained teacher is used as the teacher, while the student employs models distilled using each KD method. 
% KD method들의 매 generation step까지의 ExAccErr 평균과 model 학습 과정에서의 50 generation step까지의 ExAccErr 변화를 Figure에 depict했다. In this experiment, the teacher refers to a fixed pre-trained teacher, while the student employs models distilled using each KD method. 

When examining the ExAccErr over generation steps in Figure~\ref{fig:eb_length}, it can be observed that for most methods, the error due to exposure bias accumulates as the generation length increases, increasing ExAccErr values. In the case of GKD, the objective used in training leads the student to minimize $R(l)$. Consequently, the value becomes negative, indicating that the distribution gap between the student and the teacher approaches 0 when using a student-generated response as a pseudo-target. However, there still exists a distribution gap for the teacher's oracle response, and this means exposure bias also still exists. Nevertheless, PromptKD maintains ExAccErr values close to 0 at all generation steps, indicating that error accumulation does not occur. This demonstrates that PromptKD is the most effective in alleviating exposure bias compared to other baselines. 
% 먼저 Figure 1의 generation step에 따른 ExAccErr의 평균을 봤을 때, 대부분의 method는 generation length가 증가함에 따라 exposure bias에 의한 error가 누적되어 ExAccErr 값이 커지는 것을 볼 수 있다. GKD의 경우, 학습에 사용되는 objective에 의해 R(l)을 최소화하는 방향으로 student가 학습되어서 값이 음수가 된다. 즉, student가 생성한 response에 대해서는 objective에 의해 teacher와의 분포 차이가 0에 가까워졌지만, teacher의 oracle response에 대해선 여전히 분포 차이가 존재함을 의미한다. 그럼에도 불구하고, PromptKD는 ExAccErr 값이 모든 generation step에서 0과 유사하게 유지되며 error가 누적되지 않는다. 이를 통해 PromptKD가 다른 baseline들에 비해 가장 효과적으로 exposure bias를 해소했다는 것을 확인한다. 

Furthermore, ExAccErr is measured up to 50 generation steps in Figure~\ref{fig:eb_training_step} to focus on the early generations where errors tend to accumulate. To observe how it changes during the training process, the total training step of best checkpoint is divided by 10, and the model is saved at each time step for ExAccErr measurement. It is apparent that PromptKD, MiniLLM, and GKD, which utilize student's responses, exhibit consistently lower ExAccErr values compared to other baselines from the early stages of training. Among them, PromptKD demonstrates the most stable maintenance of ExAccErr close to 0, signifying that distilling student-friendly knowledge aids in mitigating exposure bias during training.
% 또한, Figure 2에서 볼 수 있듯, error가 누적되기 쉬운 generation 초기에 집중하기 위해 50 generation step까지의 ExAccErr를 측정했는데, 학습 과정동안 어떻게 변하는지를 확인하기 위해 최종 training step을 10으로 나누어 각 time step때 model을 저장한 뒤 측정했다. Student의 response를 사용하는 PromptKD, MINILLM, GKD에서 확실히 other baseline보다 학습 초기에 빠르게 ExAccErr 값이 작아지는 것을 볼 수 있다. 그 중 PromptKD가 가장 안정적으로 ExAccErr를 0에 가깝게 유지하는 것을 통해, student-friendly knowledge를 distill하는 것이 학습 초기에 exposure bias를 해소하는 것에 도움이 되는 것을 demonstrate한다.

\begin{table*}[!t]
\centering
\begin{tabular}{l|p{0.8\linewidth}}
\hline
\multicolumn{2}{c}{Case 1} \\
\hline
Instruction & Tell me whether these are books by black authors: I Know Why the Caged Bird Sings, Homegoing, Between the World and Me, Becoming, Beloved, The Color Purple. \\
\hline
Ground Truth & The books: I Know Why the Caged Bird Sings, Homegoing, Between the World and Me, Becoming, Beloved, and The Color Purple are all written by black authors. \\
\hline
\hline
Teacher & All of the books you mentioned are by black authors. I Know Why the Caged Bird \\
w/o Prompt & Sings and Beloved are by Maya Angelou. Homegoing is by Yaa Gyasi. Between the World and Me is by Ta-Nehisi Coates. Becoming is by Michelle Obama. The Color Purple is by Alice Walker. \\
\hline
Teacher & I Know Why the Caged Bird Sings, Homegoing, Between the World and Me, Be- \\
w/ Prompt & coming, and The Color Purple are all books by black authors. \\
\hline
Student & Yes, these are all books by black authors. \\
\hline
\hline
\multicolumn{2}{c}{Case 2} \\
\hline
Instruction & Tell me which one does not fit with the group:  Football, Volleyball, Cricket, Chess, Hockey \\
\hline
Ground Truth & All except Chess are outdoor games. \\
\hline
\hline
Teacher & All of these are sports. However, Chess and Hockey are the odd ones out. Chess is a \\
w/o Prompt & board game, and Hockey is a winter sport. \\
\hline
Teacher & All except Chess are sports that are played in teams \\
w/ Prompt & \\
\hline
Student & All of the given options are sports. The one that does not fit with the group is Chess. The other options are sports. \\
\hline
% Distilled student & Yes, these are all books by black authors. \\
% \hline
% \hline
% Instruction & Tell me which one does not fit with the group:  Football, Volleyball, Cricket, Chess, Hockey \\
% \hline
% Ground Truth & All except Chess are outdoor games. \\
% \hline
% \hline
% Teacher w/o prompt & All of these are sports. However, Chess and Hockey are the odd ones out. Chess is a board game, and Hockey is a winter sport. \\
% \hline
% Teacher w/ prompt & All except Chess are sports that are played in teams \\
% \hline
% Initial student & None of the above \\
% \hline
% Distilled student & All of the given options are sports. The one that does not fit with the group is Chess. The other options are sports. \\
% \hline
\end{tabular}
\caption{Qualitative results of generated response from the Dolly validation set with and without using prompts for the Llama-13B teacher. A teacher with a prompt generates a response more similar to that of the student.}
\label{tab:w_wo_prompt_qual_1}
%\vspace{-0.1in}
\end{table*}
% \begin{table*}[!t]
% \centering
% \begin{tabular}{l|p{0.8\linewidth}}
% % \hline
% % \multicolumn{2}{c}{Case 1} \\
% \hline
% \end{tabular}
% \caption{Another sample response from the Dolly validation set. A teacher with a prompt produces a concise sentence that is easy for the student to understand.}
% \label{tab:w_wo_prompt_qual_2}
% \end{table*}

\begin{table}[!t]
\centering
\begin{tabular}{lccc}
\hline
\multirow{2}{*}{Method} & MA & CA & Time \\
 & (GB) & (GB) & (hour) \\
\hline
\hline
SFT & 15.70 & 28.90 & 15.70 \\
KD & 40.13 & 52.82 & 20.62 \\
SeqKD & 40.13 & 52.82 & 20.13 \\
GKD & 41.99 & 56.13 & 25.37 \\
MiniLLM & \textbf{68.91} & \textbf{78.54} & \textbf{85.71} \\
PromptKD & 43.62 & 56.57 & 26.97 \\
\hline
\end{tabular}
\caption{Comparison of computational costs. Where MA denotes the maximum allocated memory on the GPU and CA denotes the maximum cached memory on the GPU. Time indicates the total training time for each method. All computational costs are calculated on 4 NVIDIA A100 80 GB (PCIe) GPUs.}
%a batch size of 2 and a gradient accumulation step of 8.
%\textbf{Training costs} Computational costs during training are denoted as follows: MA represents the maximum allocated memory on the GPU, and CA denotes the maximum cached memory on the GPU. Time indicates the total training time for each method, expressed in hours. All computational costs are calculated using a batch size of 2 and a gradient accumulation step of 8.}
\label{tab:computational_cost}
\vspace{-0.1in}
\end{table}

\paragraph{Computational cost}

To demonstrate the efficiency of PromptKD, we compare its computational cost with baselines in Table~\ref{tab:computational_cost}. OPT-13B and OPT-6.7B are used as the teacher and the student, with measurements conducted on 4 NVIDIA A100 80GB (PCIe) GPUs. From a time perspective, methods that sample the student at each iteration to create pseudo-targets take significantly more time than those that do not. In particular, MiniLLM requires a significant amount of time, primarily due to the additional use of the corpus used for pre-training, along with the complexity of calculating intricate rewards for optimization with policy gradient, unlike other methods. For the same reason, MiniLLM demands a substantial amount of memory. In contrast, PromptKD adds only a minimal amount of memory by introducing parameters equivalent to the product of prompt length and input embedding dimension. PromptKD demonstrates clear efficiency over MiniLLM and comparable costs to GKD, while significantly outperforming both in terms of performance. Therefore, PromptKD proves competitive in this regard.

\begin{table*}[!t]
\centering
\begin{tabular}{lcccc|ccc}
\hline
\multirow{2}{*}{Model} & \multirow{2}{*}{Prompt} & \multicolumn{3}{c|}{Training set (seen)} & \multicolumn{3}{c}{Validation set (unseen)} \\ 
\cline{3-5} \cline{6-8} &  & KLD w/ $S_\text{i}$ & KLD w/ $S_\text{f}$ & \multicolumn{1}{c|}{ROUGE-L} & KLD w/ $S_\text{i}$ & KLD w/ $S_\text{f}$ & ROUGE-L \\ 
\hline
\hline
\multirow{2}{*}{GPT-2} 
& \xmark & 1.7426 & 2.2896 & \textbf{96.510} & 0.9203 & 1.0631 & \textbf{29.695}\\ 
& \cmark & \textbf{1.7416} & \textbf{2.2882} & 74.659 & \textbf{0.9069} & \textbf{1.0261} & 26.893\\ 
\hline
\multirow{2}{*}{OPT} 
& \xmark & 1.2360 & 1.6180 & \textbf{89.969} & 0.7038 & 0.8302 & 31.603\\ 
& \cmark & \textbf{1.2299} & \textbf{1.6089} & 89.137 & \textbf{0.6988} & \textbf{0.8065} & \textbf{31.933}\\ 
\hline
\multirow{2}{*}{Llama} 
& \xmark & 1.3193 & 1.9413 & 96.951 & 0.7279 & 0.9335 & 35.116 \\ 
& \cmark & \textbf{1.3186} & \textbf{1.9405} & \textbf{97.095} & \textbf{0.7184} & \textbf{0.9123} & \textbf{35.168}\\ 
\hline
\end{tabular}
\caption{Quantitative comparison between the teacher with prompt and without prompt. Measurements are conducted on both the training set and the validation set. $S_\text{i}$ and $S_\text{f}$ denote the student at the beginning and end of distillation, respectively. ROUGE-L evaluates how similar the responses are to the ground truth for each dataset. For each model, the smaller KL divergence values and larger ROUGE-L scores are highlighted in \textbf{boldface}.}
\label{tab:student_friendly}
\vspace{-0.07in}
\end{table*}

\paragraph{Student-friendly knowledge}
To provide a clear interpretation of student-friendly knowledge, we investigate how the prompt modifies the teacher model. As shown in Table~\ref{tab:w_wo_prompt_qual_1}, we generate responses to a validation set that was unseen during training using both teacher models—with and without prompt—and the trained student model. The findings reveal that while the original teacher generates a complex response, the student-friendly teacher, modified by the prompt, produces a response that is similar to and easily understood by the student. Notably, despite its simplicity, this response remains accurate. 

%Furthermore, when modifying the teacher using the prompt, the quantitative verification of maintaining quality while achieving similarity to the student in responses is detailed in Appendix~\ref{sec:stu_fri_know}. 

Furthermore, akin to the training process where responses are fed into both models via teacher-forcing, we measure the KL divergence between the output of the teacher and student model in the response part. Here, the student models considered are both at the beginning and end of distillation. Additionally, we generate responses directly and evaluate their ROUGE-L score against ground truth. For the dataset, we use 1000 samples from each, specifically from the Dolly training set observed during training and the Dolly validation set unseen during training. For each model family, we use GPT-XL (1.5B), OPT-13B, and Llama-13B as the teacher models, and GPT-Large (760M), OPT-6.7B, and Llama-7B as the student models. 

Examining the KL divergence in Table~\ref{tab:student_friendly} first, it is evident that the teacher using prompts achieves a smaller KL divergence value compared to the student at the end of distillation, as encouraged by the given objective. However, this trend is also observed with the validation set. This pattern appears across all models, indicating that using prompts makes the teacher operate more like a general language model at a similar level to the student. Moreover, the teacher using prompts exhibits prediction distributions even closer to the initial student, before distillation has taken place.

When considering ROUGE-L scores, it is observed that as the model size increases, the teacher using prompts generates responses more similar to the ground truth. This suggests that with smaller models, the teacher is adversely affected by the low level of the student when training prompts to distill student-friendly knowledge. Nevertheless, the results from the Llama model indicate that the teacher becoming similar to the student's predictive distribution does not imply a decline in its instruction-following performance.

Therefore, the student-friendly knowledge distilled in PromptKD refers to knowledge transferred by a student-friendly teacher, who maintains a similar output distribution to the student for easier understanding while preserving the original generative performance. This aligns with the concept of adaptive teaching that served as the inspiration.

%\NEW{student-friendly knowledge에 대한 명쾌한 해석을 위해, prompt가 teacher model을 어떻게 변형시키는지 확인한다. prompt를 사용할때와 사용하지 않을때의 teacher model과 학습이 완료된 student model을 사용하여 학습 과정에서 보지 못한 validation set에 대한 response를 생성해 보았으며, Table 2에 이를 나타낸다. 생성 결과, 기존의 teacher는 response를 복잡하게 생성한 반면, prompt를 통해 변형된 student-friendly한 teacher는 student와 유사하면서 이해하기 쉬운 response를 생성했다. 특히, 조금 더 단순하지만 오답이 아니라는 점은 강조할만 하다. 또한, 이처럼 prompt를 통해 teacher를 변형시켰을 때 response가 student와 유사하면서 quality는 유지되는 것을 정량적으로 확인하였는데, 이에 대한 실험 세팅과 결과는 Appendix B를 참고하라. 결과적으로, 본 논문에서 adaptive teaching에 영감을 받아 정의한 student-friendly knowledge란, student와 비슷한 예측 분포를 가지면서 생성 성능은 기존처럼 유지하는 student-friendly teacher가 전수하는 knowledge이다.}
% 마지막 문장: 따라서, PromptKD에서 전수한 student-friendly knowledge란 student가 쉽게 이해할 수 있도록 student와 비슷한 예측 분포를 가지면서 생성 성능은 기존처럼 유지하는 student-friendly teacher가 전수하는 knowledge이며, 이는 영감을 받았던 adaptive teaching이란 개념과 일치한다.

\paragraph{Ablation study}
Due to the page limit, we detail an ablation study on parameter-efficient fine-tuning methods, regularization loss, prompt settings, and KL divergence in Appendix~\ref{sec:abl_study}.

% \NEW{우리는 Regularization loss, Prompt settings, KL divergence에 대한 ablation study를 Appendix A에 정리했다.}

\section{Conclusions}

In this work, we have pioneered the exploration of extracting and distilling student-friendly knowledge for generative language models. To achieve this, we have proposed a novel method called PromptKD, which leverages prompt tuning in knowledge distillation for the first time. Owing to the memory-efficient nature of prompts and the advantage of replacing full-parameter fine-tuning, particularly beneficial for larger models like LLMs, PromptKD has proven to be an efficient approach. Through extensive experiments, PromptKD has achieved state-of-the-art performance, confirming the effectiveness of student-friendly knowledge in generation tasks. Specifically, it has been revealed that this student-friendly knowledge is extracted from a modified teacher, which outputs a distribution similar to that of the student while maintaining the generation performance. Moreover, through exposure bias analysis, we have demonstrated that PromptKD successfully alleviates exposure bias throughout the training process.
% In this work, 우리는 LLM과 같은 generative language model을 대상으로 knowledge distillation 하는데 있어 student-friendly knowledge를 추출하고 전수하는 것의 효과를 처음으로 explore했다. 이를 위해 prompt tuning을 knowledge distillation에 처음으로 활용하는 novel한 PromptKD method를 제안했으며, prompt의 memory-efficient하며 model의 크기가 클수록 full-parameter fine-tuning을 대체할 수 있다는 장점 덕분에, PromptKD는 LLM에도 효과적으로 적용 가능한 방법이다. Extensive experiments 결과 SOTA의 성능을 달성함으로써 student-friendly knowledge가 generation task에서도 잘 working하는 것을 확인했으며, exposure bias analysis를 통해 학습 전반에 걸쳐 PromptKD가 exposure bias를 효과적으로 해소하는 것을 증명했다. 
% 특히, 이러한 student-friendly knowledge는 student와 유사하지만 성능이 떨어지지는 않는 teacher로부터 추출된 knowledge라는 것도 밝혀내었다.  

\section*{Limitations}

While PromptKD has achieved state-of-the-art performance by distilling student-friendly knowledge, it still has limitations in terms of its naive extraction approach. Considering that knowledge distillation (KD) research for classification tasks employs various methods to distill student-friendly knowledge, it is expected that there are alternative approaches to effectively transfer student-friendly knowledge in a generative language model. Furthermore, although PromptKD is designed for instruction-following settings based on task-specific KD, there is a need for expansion towards task-agnostic KD to make it usable during the pre-training process.
% 비록 PromptKD가 student-friendly knowledge를 전수하여 SOTA를 달성했지만, 여전히 추출하는 방식이 naive하다는 한계가 있다. classification task를 위한 KD 연구들은 다양한 방식으로 student-friendly knowledge를 distill하기 때문에, generative language model에서도 충분히 student-friendly knowledge를 전수하는 다른 방법이 있을 것으로 기대한다. 또한 PromptKD는 task-specific KD 기반의 방법론으로 instruction-following setting에서 고안되었지만, 앞으로는 pre-training 과정에서도 사용 가능하도록 task-agnostic KD로의 확장이 필요할 것이다. 

\section*{Ethics Statement}

PromptKD utilizes pre-trained models, exposing it to risks similar to those highlighted by \citet{weidinger2021ethical,bommasani2021opportunities}, regarding the vulnerability of pre-trained language models to ethical and social risks. Additionally, \citet{hooker2020characterising} mentions that the process of model compression can introduce biases. However, since most model compression studies leverage pre-trained models, these issues are general risks and not specific to PromptKD. Nevertheless, these risks should be addressed in the future through advanced pre-training objectives and dataset collection methods~\citep{lee-etal-2023-kosbi}.
% 우리 PromptKD는 pre-trained model을 사용하기 때문에, Bommasani나 Weiginger가 언급한 것처럼 pre-trained language model이 ethical and social risk에 취약하다는 것과 비슷한 risk에 노출된다. 또한, Hooker가 언급한 것과 같이 model compression 과정이 biases를 이끌 수 있다. 그러나, 대부분의 model compression 연구들은 pre-trained model을 활용하기 때문에, 이러한 부분들은 일반적인 risk이며 PromptKD에서 기인하지 않는다. 또한 이러한 risk들은 고도화된 pre-training objective나 dataset collection 등으로 미래엔 해결되어야한다. 

\section*{Acknowledgements}
This work was supported by Institute for Information \& communications Technology Promotion (IITP) grant funded by the Korea government (MSIT) (No.RS-2019-II190075, Artificial Intelligence Graduate School Program(KAIST), No. RS-2024-00457882, AI Research Hub Project).
%No.2021-0-02068, Artificial Intelligence Innovation Hub, No.2022-0-00713, Meta-learning applicable to real-world problems, No.2022-0-00984, Development of Artificial Intelligence Technology for Personalized Plug-and-Play Explanation and Verification of Explanation)
%and the National Research Foundation of Korea (NRF) grants (No.2018R1A5A1059921) 

% Bibliography entries for the entire Anthology, followed by custom entries
\bibliography{anthology,custom}

\begin{thebibliography}{57}
\expandafter\ifx\csname natexlab\endcsname\relax\def\natexlab#1{#1}\fi

\bibitem[{Agarwal et~al.(2024)Agarwal, Vieillard, Zhou, Stanczyk, Garea, Geist, and Bachem}]{gkd}
Rishabh Agarwal, Nino Vieillard, Yongchao Zhou, Piotr Stanczyk, Sabela~Ramos Garea, Matthieu Geist, and Olivier Bachem. 2024.
\newblock \href {https://openreview.net/forum?id=3zKtaqxLhW} {Generalized knowledge distillation for auto-regressive language models}.
\newblock In \emph{The Twelfth International Conference on Learning Representations}.

\bibitem[{Arora et~al.(2022)Arora, El~Asri, Bahuleyan, and Cheung}]{arora-etal-2022-exposure}
Kushal Arora, Layla El~Asri, Hareesh Bahuleyan, and Jackie Cheung. 2022.
\newblock \href {https://doi.org/10.18653/v1/2022.findings-acl.58} {Why exposure bias matters: An imitation learning perspective of error accumulation in language generation}.
\newblock In \emph{Findings of the Association for Computational Linguistics: ACL 2022}, pages 700--710, Dublin, Ireland. Association for Computational Linguistics.

\bibitem[{Bommasani et~al.(2021)Bommasani, Hudson, Adeli, Altman, Arora, von Arx, Bernstein, Bohg, Bosselut, Brunskill et~al.}]{bommasani2021opportunities}
Rishi Bommasani, Drew~A Hudson, Ehsan Adeli, Russ Altman, Simran Arora, Sydney von Arx, Michael~S Bernstein, Jeannette Bohg, Antoine Bosselut, Emma Brunskill, et~al. 2021.
\newblock On the opportunities and risks of foundation models.
\newblock \emph{arXiv preprint arXiv:2108.07258}.

\bibitem[{Brown et~al.(2020)Brown, Mann, Ryder, Subbiah, Kaplan, Dhariwal, Neelakantan, Shyam, Sastry, Askell et~al.}]{brown2020language}
Tom Brown, Benjamin Mann, Nick Ryder, Melanie Subbiah, Jared~D Kaplan, Prafulla Dhariwal, Arvind Neelakantan, Pranav Shyam, Girish Sastry, Amanda Askell, et~al. 2020.
\newblock Language models are few-shot learners.
\newblock \emph{Advances in neural information processing systems}, 33:1877--1901.

\bibitem[{Calderon et~al.(2023)Calderon, Mukherjee, Reichart, and Kantor}]{calderon-etal-2023-systematic}
Nitay Calderon, Subhabrata Mukherjee, Roi Reichart, and Amir Kantor. 2023.
\newblock \href {https://doi.org/10.18653/v1/2023.acl-long.818} {A systematic study of knowledge distillation for natural language generation with pseudo-target training}.
\newblock In \emph{Proceedings of the 61st Annual Meeting of the Association for Computational Linguistics (Volume 1: Long Papers)}, pages 14632--14659, Toronto, Canada. Association for Computational Linguistics.

\bibitem[{Chiang et~al.(2023)Chiang, Li, Lin, Sheng, Wu, Zhang, Zheng, Zhuang, Zhuang, Gonzalez et~al.}]{chiang2023vicuna}
Wei-Lin Chiang, Zhuohan Li, Zi~Lin, Ying Sheng, Zhanghao Wu, Hao Zhang, Lianmin Zheng, Siyuan Zhuang, Yonghao Zhuang, Joseph~E Gonzalez, et~al. 2023.
\newblock Vicuna: An open-source chatbot impressing gpt-4 with 90\%* chatgpt quality.
\newblock \emph{See https://vicuna. lmsys. org (accessed 14 April 2023)}.

\bibitem[{Cho and Hariharan(2019)}]{cho2019efficacy}
Jang~Hyun Cho and Bharath Hariharan. 2019.
\newblock On the efficacy of knowledge distillation.
\newblock In \emph{Proceedings of the IEEE/CVF international conference on computer vision}, pages 4794--4802.

\bibitem[{Conover et~al.(2023)Conover, Hayes, Mathur, Xie, Wan, Shah, Ghodsi, Wendell, Zaharia, and Xin}]{databrick2023dollyv2}
Mike Conover, Matt Hayes, Ankit Mathur, Jianwei Xie, Jun Wan, Sam Shah, Ali Ghodsi, Patrick Wendell, Matei Zaharia, and Reynold Xin. 2023.
\newblock \href {https://www.databricks.com/blog/2023/04/12/dolly-first-open-commercially-viable-instruction-tuned-llm} {Free dolly: Introducing the world's first truly open instruction-tuned llm}.

\bibitem[{Devlin et~al.(2019)Devlin, Chang, Lee, and Toutanova}]{devlin-etal-2019-bert}
Jacob Devlin, Ming-Wei Chang, Kenton Lee, and Kristina Toutanova. 2019.
\newblock \href {https://doi.org/10.18653/v1/N19-1423} {{BERT}: Pre-training of deep bidirectional transformers for language understanding}.
\newblock In \emph{Proceedings of the 2019 Conference of the North {A}merican Chapter of the Association for Computational Linguistics: Human Language Technologies, Volume 1 (Long and Short Papers)}, pages 4171--4186, Minneapolis, Minnesota. Association for Computational Linguistics.

\bibitem[{Gao et~al.(2021)Gao, Fisch, and Chen}]{gao-etal-2021-making}
Tianyu Gao, Adam Fisch, and Danqi Chen. 2021.
\newblock \href {https://doi.org/10.18653/v1/2021.acl-long.295} {Making pre-trained language models better few-shot learners}.
\newblock In \emph{Proceedings of the 59th Annual Meeting of the Association for Computational Linguistics and the 11th International Joint Conference on Natural Language Processing (Volume 1: Long Papers)}, pages 3816--3830, Online. Association for Computational Linguistics.

\bibitem[{Gu et~al.(2024)Gu, Dong, Wei, and Huang}]{minillm}
Yuxian Gu, Li~Dong, Furu Wei, and Minlie Huang. 2024.
\newblock \href {https://openreview.net/forum?id=5h0qf7IBZZ} {Mini{LLM}: Knowledge distillation of large language models}.
\newblock In \emph{The Twelfth International Conference on Learning Representations}.

\bibitem[{Hambardzumyan et~al.(2021)Hambardzumyan, Khachatrian, and May}]{hambardzumyan-etal-2021-warp}
Karen Hambardzumyan, Hrant Khachatrian, and Jonathan May. 2021.
\newblock \href {https://doi.org/10.18653/v1/2021.acl-long.381} {{WARP}: {W}ord-level {A}dversarial {R}e{P}rogramming}.
\newblock In \emph{Proceedings of the 59th Annual Meeting of the Association for Computational Linguistics and the 11th International Joint Conference on Natural Language Processing (Volume 1: Long Papers)}, pages 4921--4933, Online. Association for Computational Linguistics.

\bibitem[{Hinton et~al.(2015)Hinton, Vinyals, and Dean}]{hinton2015distilling}
Geoffrey Hinton, Oriol Vinyals, and Jeff Dean. 2015.
\newblock Distilling the knowledge in a neural network.
\newblock \emph{arXiv preprint arXiv:1503.02531}.

\bibitem[{Ho et~al.(2023)Ho, Schmid, and Yun}]{ho-etal-2023-large}
Namgyu Ho, Laura Schmid, and Se-Young Yun. 2023.
\newblock \href {https://doi.org/10.18653/v1/2023.acl-long.830} {Large language models are reasoning teachers}.
\newblock In \emph{Proceedings of the 61st Annual Meeting of the Association for Computational Linguistics (Volume 1: Long Papers)}, pages 14852--14882, Toronto, Canada. Association for Computational Linguistics.

\bibitem[{Honovich et~al.(2023)Honovich, Scialom, Levy, and Schick}]{honovich-etal-2023-unnatural}
Or~Honovich, Thomas Scialom, Omer Levy, and Timo Schick. 2023.
\newblock \href {https://doi.org/10.18653/v1/2023.acl-long.806} {Unnatural instructions: Tuning language models with (almost) no human labor}.
\newblock In \emph{Proceedings of the 61st Annual Meeting of the Association for Computational Linguistics (Volume 1: Long Papers)}, pages 14409--14428, Toronto, Canada. Association for Computational Linguistics.

\bibitem[{Hooker et~al.(2020)Hooker, Moorosi, Clark, Bengio, and Denton}]{hooker2020characterising}
Sara Hooker, Nyalleng Moorosi, Gregory Clark, Samy Bengio, and Emily Denton. 2020.
\newblock Characterising bias in compressed models.
\newblock \emph{arXiv preprint arXiv:2010.03058}.

\bibitem[{Hou et~al.(2022)Hou, Dong, Wang, Li, and Che}]{hou-etal-2022-metaprompting}
Yutai Hou, Hongyuan Dong, Xinghao Wang, Bohan Li, and Wanxiang Che. 2022.
\newblock \href {https://aclanthology.org/2022.coling-1.287} {{M}eta{P}rompting: Learning to learn better prompts}.
\newblock In \emph{Proceedings of the 29th International Conference on Computational Linguistics}, pages 3251--3262, Gyeongju, Republic of Korea. International Committee on Computational Linguistics.

\bibitem[{Hsieh et~al.(2023)Hsieh, Li, Yeh, Nakhost, Fujii, Ratner, Krishna, Lee, and Pfister}]{hsieh-etal-2023-distilling}
Cheng-Yu Hsieh, Chun-Liang Li, Chih-kuan Yeh, Hootan Nakhost, Yasuhisa Fujii, Alex Ratner, Ranjay Krishna, Chen-Yu Lee, and Tomas Pfister. 2023.
\newblock \href {https://doi.org/10.18653/v1/2023.findings-acl.507} {Distilling step-by-step! outperforming larger language models with less training data and smaller model sizes}.
\newblock In \emph{Findings of the Association for Computational Linguistics: ACL 2023}, pages 8003--8017, Toronto, Canada. Association for Computational Linguistics.

\bibitem[{Hu et~al.(2022)Hu, yelong shen, Wallis, Allen-Zhu, Li, Wang, Wang, and Chen}]{hu2022lora}
Edward~J Hu, yelong shen, Phillip Wallis, Zeyuan Allen-Zhu, Yuanzhi Li, Shean Wang, Lu~Wang, and Weizhu Chen. 2022.
\newblock \href {https://openreview.net/forum?id=nZeVKeeFYf9} {Lo{RA}: Low-rank adaptation of large language models}.
\newblock In \emph{International Conference on Learning Representations}.

\bibitem[{Jiang et~al.(2020)Jiang, Xu, Araki, and Neubig}]{jiang-etal-2020-know}
Zhengbao Jiang, Frank~F. Xu, Jun Araki, and Graham Neubig. 2020.
\newblock \href {https://doi.org/10.1162/tacl_a_00324} {How can we know what language models know?}
\newblock \emph{Transactions of the Association for Computational Linguistics}, 8:423--438.

\bibitem[{Jiao et~al.(2020)Jiao, Yin, Shang, Jiang, Chen, Li, Wang, and Liu}]{jiao-etal-2020-tinybert}
Xiaoqi Jiao, Yichun Yin, Lifeng Shang, Xin Jiang, Xiao Chen, Linlin Li, Fang Wang, and Qun Liu. 2020.
\newblock \href {https://doi.org/10.18653/v1/2020.findings-emnlp.372} {{T}iny{BERT}: Distilling {BERT} for natural language understanding}.
\newblock In \emph{Findings of the Association for Computational Linguistics: EMNLP 2020}, pages 4163--4174, Online. Association for Computational Linguistics.

\bibitem[{Kim and Rush(2016)}]{kim-rush-2016-sequence}
Yoon Kim and Alexander~M. Rush. 2016.
\newblock \href {https://doi.org/10.18653/v1/D16-1139} {Sequence-level knowledge distillation}.
\newblock In \emph{Proceedings of the 2016 Conference on Empirical Methods in Natural Language Processing}, pages 1317--1327, Austin, Texas. Association for Computational Linguistics.

\bibitem[{Lee et~al.(2023)Lee, Hong, Park, Kim, Kim, and Ha}]{lee-etal-2023-kosbi}
Hwaran Lee, Seokhee Hong, Joonsuk Park, Takyoung Kim, Gunhee Kim, and Jung-woo Ha. 2023.
\newblock \href {https://doi.org/10.18653/v1/2023.acl-industry.21} {{K}o{SBI}: A dataset for mitigating social bias risks towards safer large language model applications}.
\newblock In \emph{Proceedings of the 61st Annual Meeting of the Association for Computational Linguistics (Volume 5: Industry Track)}, pages 208--224, Toronto, Canada. Association for Computational Linguistics.

\bibitem[{Lester et~al.(2021)Lester, Al-Rfou, and Constant}]{lester-etal-2021-power}
Brian Lester, Rami Al-Rfou, and Noah Constant. 2021.
\newblock \href {https://doi.org/10.18653/v1/2021.emnlp-main.243} {The power of scale for parameter-efficient prompt tuning}.
\newblock In \emph{Proceedings of the 2021 Conference on Empirical Methods in Natural Language Processing}, pages 3045--3059, Online and Punta Cana, Dominican Republic. Association for Computational Linguistics.

\bibitem[{Li and Liang(2021)}]{li-liang-2021-prefix}
Xiang~Lisa Li and Percy Liang. 2021.
\newblock \href {https://doi.org/10.18653/v1/2021.acl-long.353} {Prefix-tuning: Optimizing continuous prompts for generation}.
\newblock In \emph{Proceedings of the 59th Annual Meeting of the Association for Computational Linguistics and the 11th International Joint Conference on Natural Language Processing (Volume 1: Long Papers)}, pages 4582--4597, Online. Association for Computational Linguistics.

\bibitem[{Liang et~al.(2023{\natexlab{a}})Liang, Jiang, Li, Tang, Yin, and Zhao}]{liang2023homodistil}
Chen Liang, Haoming Jiang, Zheng Li, Xianfeng Tang, Bing Yin, and Tuo Zhao. 2023{\natexlab{a}}.
\newblock \href {https://openreview.net/forum?id=D7srTrGhAs} {Homodistil: Homotopic task-agnostic distillation of pre-trained transformers}.
\newblock In \emph{The Eleventh International Conference on Learning Representations}.

\bibitem[{Liang et~al.(2023{\natexlab{b}})Liang, Zuo, Zhang, He, Chen, and Zhao}]{liang2023less}
Chen Liang, Simiao Zuo, Qingru Zhang, Pengcheng He, Weizhu Chen, and Tuo Zhao. 2023{\natexlab{b}}.
\newblock Less is more: Task-aware layer-wise distillation for language model compression.
\newblock In \emph{International Conference on Machine Learning}, pages 20852--20867. PMLR.

\bibitem[{Lin(2004)}]{lin-2004-rouge}
Chin-Yew Lin. 2004.
\newblock \href {https://aclanthology.org/W04-1013} {{ROUGE}: A package for automatic evaluation of summaries}.
\newblock In \emph{Text Summarization Branches Out}, pages 74--81, Barcelona, Spain. Association for Computational Linguistics.

\bibitem[{Loshchilov and Hutter(2019)}]{loshchilov2018decoupled}
Ilya Loshchilov and Frank Hutter. 2019.
\newblock \href {https://openreview.net/forum?id=Bkg6RiCqY7} {Decoupled weight decay regularization}.
\newblock In \emph{International Conference on Learning Representations}.

\bibitem[{Ma et~al.(2023)Ma, Fang, and Wang}]{ma2023llmpruner}
Xinyin Ma, Gongfan Fang, and Xinchao Wang. 2023.
\newblock \href {https://doi.org/10.48550/arXiv.2305.11627} {Llm-pruner: On the structural pruning of large language models}.
\newblock \emph{CoRR}, abs/2305.11627.

\bibitem[{Ma et~al.(2022)Ma, Wang, Fang, Shen, and Lu}]{ijcai2022p596}
Xinyin Ma, Xinchao Wang, Gongfan Fang, Yongliang Shen, and Weiming Lu. 2022.
\newblock \href {https://doi.org/10.24963/ijcai.2022/596} {Prompting to distill: Boosting data-free knowledge distillation via reinforced prompt}.
\newblock In \emph{Proceedings of the Thirty-First International Joint Conference on Artificial Intelligence, {IJCAI-22}}, pages 4296--4302. International Joint Conferences on Artificial Intelligence Organization.
\newblock Main Track.

\bibitem[{Nowozin et~al.(2016)Nowozin, Cseke, and Tomioka}]{nowozin2016f}
Sebastian Nowozin, Botond Cseke, and Ryota Tomioka. 2016.
\newblock f-gan: Training generative neural samplers using variational divergence minimization.
\newblock \emph{Advances in neural information processing systems}, 29.

\bibitem[{Ouyang et~al.(2022)Ouyang, Wu, Jiang, Almeida, Wainwright, Mishkin, Zhang, Agarwal, Slama, Ray et~al.}]{ouyang2022training}
Long Ouyang, Jeffrey Wu, Xu~Jiang, Diogo Almeida, Carroll Wainwright, Pamela Mishkin, Chong Zhang, Sandhini Agarwal, Katarina Slama, Alex Ray, et~al. 2022.
\newblock Training language models to follow instructions with human feedback.
\newblock \emph{Advances in Neural Information Processing Systems}, 35:27730--27744.

\bibitem[{Park et~al.(2021{\natexlab{a}})Park, Cha, Kim, Han et~al.}]{park2021learning}
Dae~Young Park, Moon-Hyun Cha, Daesin Kim, Bohyung Han, et~al. 2021{\natexlab{a}}.
\newblock Learning student-friendly teacher networks for knowledge distillation.
\newblock \emph{Advances in neural information processing systems}, 34:13292--13303.

\bibitem[{Park et~al.(2021{\natexlab{b}})Park, Kim, and Yang}]{park-etal-2021-distilling}
Geondo Park, Gyeongman Kim, and Eunho Yang. 2021{\natexlab{b}}.
\newblock \href {https://doi.org/10.18653/v1/2021.emnlp-main.30} {Distilling linguistic context for language model compression}.
\newblock In \emph{Proceedings of the 2021 Conference on Empirical Methods in Natural Language Processing}, pages 364--378, Online and Punta Cana, Dominican Republic. Association for Computational Linguistics.

\bibitem[{Radford et~al.(2019)Radford, Wu, Child, Luan, Amodei, Sutskever et~al.}]{radford2019language}
Alec Radford, Jeffrey Wu, Rewon Child, David Luan, Dario Amodei, Ilya Sutskever, et~al. 2019.
\newblock Language models are unsupervised multitask learners.
\newblock \emph{OpenAI blog}, 1(8):9.

\bibitem[{Ren et~al.(2023)Ren, Zhong, Shi, Zhu, Yuan, and Li}]{ren-etal-2023-tailoring}
Yuxin Ren, Zihan Zhong, Xingjian Shi, Yi~Zhu, Chun Yuan, and Mu~Li. 2023.
\newblock \href {https://doi.org/10.18653/v1/2023.acl-long.111} {Tailoring instructions to student{'}s learning levels boosts knowledge distillation}.
\newblock In \emph{Proceedings of the 61st Annual Meeting of the Association for Computational Linguistics (Volume 1: Long Papers)}, pages 1990--2006, Toronto, Canada. Association for Computational Linguistics.

\bibitem[{Sanh et~al.(2019)Sanh, Debut, Chaumond, and Wolf}]{sanh2019distilbert}
Victor Sanh, Lysandre Debut, Julien Chaumond, and Thomas Wolf. 2019.
\newblock Distilbert, a distilled version of bert: smaller, faster, cheaper and lighter.
\newblock \emph{arXiv preprint arXiv:1910.01108}.

\bibitem[{Schick and Sch{\"u}tze(2021)}]{schick-schutze-2021-exploiting}
Timo Schick and Hinrich Sch{\"u}tze. 2021.
\newblock \href {https://doi.org/10.18653/v1/2021.eacl-main.20} {Exploiting cloze-questions for few-shot text classification and natural language inference}.
\newblock In \emph{Proceedings of the 16th Conference of the European Chapter of the Association for Computational Linguistics: Main Volume}, pages 255--269, Online. Association for Computational Linguistics.

\bibitem[{Shin et~al.(2020)Shin, Razeghi, Logan~IV, Wallace, and Singh}]{shin-etal-2020-autoprompt}
Taylor Shin, Yasaman Razeghi, Robert~L. Logan~IV, Eric Wallace, and Sameer Singh. 2020.
\newblock \href {https://doi.org/10.18653/v1/2020.emnlp-main.346} {{A}uto{P}rompt: {E}liciting {K}nowledge from {L}anguage {M}odels with {A}utomatically {G}enerated {P}rompts}.
\newblock In \emph{Proceedings of the 2020 Conference on Empirical Methods in Natural Language Processing (EMNLP)}, pages 4222--4235, Online. Association for Computational Linguistics.

\bibitem[{Song et~al.(2020)Song, Sun, Tan, Qin, Lu, Liu, and Liu}]{song2020lightpaff}
Kaitao Song, Hao Sun, Xu~Tan, Tao Qin, Jianfeng Lu, Hongzhi Liu, and Tie-Yan Liu. 2020.
\newblock \href {https://openreview.net/forum?id=B1xv9pEKDS} {Light{\{}paff{\}}: A two-stage distillation framework for pre-training and fine-tuning}.

\bibitem[{Sun et~al.(2019)Sun, Cheng, Gan, and Liu}]{sun-etal-2019-patient}
Siqi Sun, Yu~Cheng, Zhe Gan, and Jingjing Liu. 2019.
\newblock \href {https://doi.org/10.18653/v1/D19-1441} {Patient knowledge distillation for {BERT} model compression}.
\newblock In \emph{Proceedings of the 2019 Conference on Empirical Methods in Natural Language Processing and the 9th International Joint Conference on Natural Language Processing (EMNLP-IJCNLP)}, pages 4323--4332, Hong Kong, China. Association for Computational Linguistics.

\bibitem[{Tao et~al.(2022)Tao, Hou, Zhang, Shang, Jiang, Liu, Luo, and Wong}]{tao-etal-2022-compression}
Chaofan Tao, Lu~Hou, Wei Zhang, Lifeng Shang, Xin Jiang, Qun Liu, Ping Luo, and Ngai Wong. 2022.
\newblock \href {https://doi.org/10.18653/v1/2022.acl-long.331} {Compression of generative pre-trained language models via quantization}.
\newblock In \emph{Proceedings of the 60th Annual Meeting of the Association for Computational Linguistics (Volume 1: Long Papers)}, pages 4821--4836, Dublin, Ireland. Association for Computational Linguistics.

\bibitem[{Taori et~al.(2023)Taori, Gulrajani, Zhang, Dubois, Li, Guestrin, Liang, and Hashimoto}]{alpaca}
Rohan Taori, Ishaan Gulrajani, Tianyi Zhang, Yann Dubois, Xuechen Li, Carlos Guestrin, Percy Liang, and Tatsunori~B. Hashimoto. 2023.
\newblock Stanford alpaca: An instruction-following llama model.
\newblock \url{https://github.com/tatsu-lab/stanford_alpaca}.

\bibitem[{Touvron et~al.(2023{\natexlab{a}})Touvron, Lavril, Izacard, Martinet, Lachaux, Lacroix, Rozi{\`e}re, Goyal, Hambro, Azhar et~al.}]{touvron2023llama1}
Hugo Touvron, Thibaut Lavril, Gautier Izacard, Xavier Martinet, Marie-Anne Lachaux, Timoth{\'e}e Lacroix, Baptiste Rozi{\`e}re, Naman Goyal, Eric Hambro, Faisal Azhar, et~al. 2023{\natexlab{a}}.
\newblock Llama: Open and efficient foundation language models.
\newblock \emph{arXiv preprint arXiv:2302.13971}.

\bibitem[{Touvron et~al.(2023{\natexlab{b}})Touvron, Martin, Stone, Albert, Almahairi, Babaei, Bashlykov, Batra, Bhargava, Bhosale et~al.}]{touvron2023llama}
Hugo Touvron, Louis Martin, Kevin Stone, Peter Albert, Amjad Almahairi, Yasmine Babaei, Nikolay Bashlykov, Soumya Batra, Prajjwal Bhargava, Shruti Bhosale, et~al. 2023{\natexlab{b}}.
\newblock Llama 2: Open foundation and fine-tuned chat models.
\newblock \emph{arXiv preprint arXiv:2307.09288}.

\bibitem[{Wang et~al.(2020)Wang, Wei, Dong, Bao, Yang, and Zhou}]{wang2020minilm}
Wenhui Wang, Furu Wei, Li~Dong, Hangbo Bao, Nan Yang, and Ming Zhou. 2020.
\newblock Minilm: Deep self-attention distillation for task-agnostic compression of pre-trained transformers.
\newblock \emph{Advances in Neural Information Processing Systems}, 33:5776--5788.

\bibitem[{Wang et~al.(2023)Wang, Kordi, Mishra, Liu, Smith, Khashabi, and Hajishirzi}]{wang-etal-2023-self-instruct}
Yizhong Wang, Yeganeh Kordi, Swaroop Mishra, Alisa Liu, Noah~A. Smith, Daniel Khashabi, and Hannaneh Hajishirzi. 2023.
\newblock \href {https://doi.org/10.18653/v1/2023.acl-long.754} {Self-instruct: Aligning language models with self-generated instructions}.
\newblock In \emph{Proceedings of the 61st Annual Meeting of the Association for Computational Linguistics (Volume 1: Long Papers)}, pages 13484--13508, Toronto, Canada. Association for Computational Linguistics.

\bibitem[{Wang et~al.(2022)Wang, Mishra, Alipoormolabashi, Kordi, Mirzaei, Naik, Ashok, Dhanasekaran, Arunkumar, Stap, Pathak, Karamanolakis, Lai, Purohit, Mondal, Anderson, Kuznia, Doshi, Pal, Patel, Moradshahi, Parmar, Purohit, Varshney, Kaza, Verma, Puri, Karia, Doshi, Sampat, Mishra, Reddy~A, Patro, Dixit, and Shen}]{wang-etal-2022-super}
Yizhong Wang, Swaroop Mishra, Pegah Alipoormolabashi, Yeganeh Kordi, Amirreza Mirzaei, Atharva Naik, Arjun Ashok, Arut~Selvan Dhanasekaran, Anjana Arunkumar, David Stap, Eshaan Pathak, Giannis Karamanolakis, Haizhi Lai, Ishan Purohit, Ishani Mondal, Jacob Anderson, Kirby Kuznia, Krima Doshi, Kuntal~Kumar Pal, Maitreya Patel, Mehrad Moradshahi, Mihir Parmar, Mirali Purohit, Neeraj Varshney, Phani~Rohitha Kaza, Pulkit Verma, Ravsehaj~Singh Puri, Rushang Karia, Savan Doshi, Shailaja~Keyur Sampat, Siddhartha Mishra, Sujan Reddy~A, Sumanta Patro, Tanay Dixit, and Xudong Shen. 2022.
\newblock \href {https://doi.org/10.18653/v1/2022.emnlp-main.340} {Super-{N}atural{I}nstructions: Generalization via declarative instructions on 1600+ {NLP} tasks}.
\newblock In \emph{Proceedings of the 2022 Conference on Empirical Methods in Natural Language Processing}, pages 5085--5109, Abu Dhabi, United Arab Emirates. Association for Computational Linguistics.

\bibitem[{Wei et~al.(2022)Wei, Tay, Bommasani, Raffel, Zoph, Borgeaud, Yogatama, Bosma, Zhou, Metzler et~al.}]{wei2022emergent}
Jason Wei, Yi~Tay, Rishi Bommasani, Colin Raffel, Barret Zoph, Sebastian Borgeaud, Dani Yogatama, Maarten Bosma, Denny Zhou, Donald Metzler, et~al. 2022.
\newblock Emergent abilities of large language models.
\newblock \emph{arXiv preprint arXiv:2206.07682}.

\bibitem[{Weidinger et~al.(2021)Weidinger, Mellor, Rauh, Griffin, Uesato, Huang, Cheng, Glaese, Balle, Kasirzadeh et~al.}]{weidinger2021ethical}
Laura Weidinger, John Mellor, Maribeth Rauh, Conor Griffin, Jonathan Uesato, Po-Sen Huang, Myra Cheng, Mia Glaese, Borja Balle, Atoosa Kasirzadeh, et~al. 2021.
\newblock Ethical and social risks of harm from language models.
\newblock \emph{arXiv preprint arXiv:2112.04359}.

\bibitem[{Yang et~al.(2022)Yang, Zhang, and Song}]{yang-etal-2022-sparse}
Yi~Yang, Chen Zhang, and Dawei Song. 2022.
\newblock \href {https://doi.org/10.18653/v1/2022.emnlp-main.258} {Sparse teachers can be dense with knowledge}.
\newblock In \emph{Proceedings of the 2022 Conference on Empirical Methods in Natural Language Processing}, pages 3904--3915, Abu Dhabi, United Arab Emirates. Association for Computational Linguistics.

\bibitem[{Zhang et~al.(2022)Zhang, Roller, Goyal, Artetxe, Chen, Chen, Dewan, Diab, Li, Lin et~al.}]{zhang2022opt}
Susan Zhang, Stephen Roller, Naman Goyal, Mikel Artetxe, Moya Chen, Shuohui Chen, Christopher Dewan, Mona Diab, Xian Li, Xi~Victoria Lin, et~al. 2022.
\newblock Opt: Open pre-trained transformer language models.
\newblock \emph{arXiv preprint arXiv:2205.01068}.

\bibitem[{Zhang et~al.(2019)Zhang, Feng, Meng, You, and Liu}]{zhang-etal-2019-bridging}
Wen Zhang, Yang Feng, Fandong Meng, Di~You, and Qun Liu. 2019.
\newblock \href {https://doi.org/10.18653/v1/P19-1426} {Bridging the gap between training and inference for neural machine translation}.
\newblock In \emph{Proceedings of the 57th Annual Meeting of the Association for Computational Linguistics}, pages 4334--4343, Florence, Italy. Association for Computational Linguistics.

\bibitem[{Zheng et~al.(2024)Zheng, Chiang, Sheng, Zhuang, Wu, Zhuang, Lin, Li, Li, Xing et~al.}]{zheng2024judging}
Lianmin Zheng, Wei-Lin Chiang, Ying Sheng, Siyuan Zhuang, Zhanghao Wu, Yonghao Zhuang, Zi~Lin, Zhuohan Li, Dacheng Li, Eric Xing, et~al. 2024.
\newblock Judging llm-as-a-judge with mt-bench and chatbot arena.
\newblock \emph{Advances in Neural Information Processing Systems}, 36.

\bibitem[{Zhong et~al.(2021)Zhong, Friedman, and Chen}]{zhong-etal-2021-factual}
Zexuan Zhong, Dan Friedman, and Danqi Chen. 2021.
\newblock \href {https://doi.org/10.18653/v1/2021.naacl-main.398} {Factual probing is [{MASK}]: Learning vs. learning to recall}.
\newblock In \emph{Proceedings of the 2021 Conference of the North American Chapter of the Association for Computational Linguistics: Human Language Technologies}, pages 5017--5033, Online. Association for Computational Linguistics.

\bibitem[{Zhou et~al.(2022)Zhou, Xu, and McAuley}]{zhou-etal-2022-bert}
Wangchunshu Zhou, Canwen Xu, and Julian McAuley. 2022.
\newblock \href {https://doi.org/10.18653/v1/2022.acl-long.485} {{BERT} learns to teach: Knowledge distillation with meta learning}.
\newblock In \emph{Proceedings of the 60th Annual Meeting of the Association for Computational Linguistics (Volume 1: Long Papers)}, pages 7037--7049, Dublin, Ireland. Association for Computational Linguistics.

\end{thebibliography}
% Custom bibliography entries only
%\bibliography{custom}

\appendix

% \section{Example Appendix}
% \label{sec:appendix}

\section{Training Details}
\label{sec:training_detail}

In our study, we employ the AdamW~\citep{loshchilov2018decoupled} optimizer for training, with batch sizes of 32 for GPT-2 Base and 8 for GPT-2 Medium and Large. The learning rates of prompt and student are set at 5e-5 for Base, 1e-5 for Medium, and 5e-6 for Large. In both the Llama and OPT model families, we set the batch size to 64 and the learning rates of prompt and student to 5e-6. For the generation, we sample with top-k and top-p parameters at 0 and 1.0, respectively, and use a fixed temperature of 1.0. Training and generation phases both have a maximum sequence length of 512 and a maximum prompt length of 256. Please note that we pre-process each instruction-following dataset by converting the instruction-response pairs into a standardized sentence structure, as shown in Table~\ref{tab:wrapper}. For the reproducibility of our PromptKD, we will make both the code and the checkpoints public.

\begin{table}[!t]
\centering
\begin{tabular}{p{0.9\linewidth}}
\hline
Prompt Format \\
\hline
Below is an instruction that describes a task. \\
Write a response that appropriately completes the request. \\
\#\#\# Instruction: \\ 
\{instruction\}  \\
\#\#\# Input: \\
\{input\} \\
\#\#\# Response: \\ 
\hline
\end{tabular}
\caption{Prompt format used for training and evaluation.}
\label{tab:wrapper}
\end{table}

\begin{table}[!t]
\centering
\begin{tabular}{lccc}
\hline
Method & Dolly & SelfInst & Vicuna \\ 
\hline
\hline
GKD & 68.83 & 63.87 & 66.68 \\
MiniLLM & 71.39 & 66.96 & 67.78 \\
PromptKD & \textbf{72.12} & \textbf{67.22} & \textbf{68.01} \\
\hline
\end{tabular}
\caption{Evaluation results with GPT-4 feedback scores.}
\label{tab:gpt4_score_result}
\end{table}

\section{Qualitative Results}
\label{sec:qual_res}
For the qualitative results, we present samples generated by student models trained using various methods. The samples are drawn from the S-NI dataset and utilize GPT-2 XL as the teacher model, with GPT-2 Large employed as the student model. Results are shown in Table~\ref{tab:quant}. Additionally, the generation results obtained using the Llama model are summarized in Table~\ref{tab:quant_llama}.

\begin{table*}[!t]
\centering
\begin{tabular}{l|p{0.8\linewidth}}
\hline
\multicolumn{2}{c}{Case 1} \\
\hline
Instruction & Convert a disfluent question to a proper question. A disfluent question is a question that has some interruptions in it while framing. A proper question is the correct form of the question without any disfluency. \\
\hline
Input & What do no I meant Who do some tax agents want to make an impression on during an arrest? \\
\hline
Ground Truth & Who do some tax agents want to make an impression on during an arrest? \\
\hline
\hline
SFT & Some tax agents want to make an impression on during an arrest. \\
\hline
KD & The tax agents want to make an impression on during an arrest. \\
\hline
SeqKD & No I meant to ask who are some tax agents that want to make an impression on an arrest. \\
\hline
GKD & No I mean that there are no tax agents who want to make an impression on the arrest. \\
\hline
MiniLLM & A disfluent question is one that has some interruptions while framing. A proper questions is the correct form of a question without any disfluidity. \\
\hline
PromptKD & Who do some tax agents want make an impression on during a arrest? \\
\hline
\hline
\multicolumn{2}{c}{Case 2} \\
\hline
Instruction & Given an entity, a before event, an after event, and an attribute related to the entity, generate a sentence as output. Your sentence should show the changes in the attribute of the entity. \\
\hline
Input & entity: rope, before: separate from ball, after: through slot in ball, attr: location \\
\hline
Ground Truth & The location of the rope was separate from the ball before and through a slot in the ball afterwards. \\
\hline
\hline
SFT & Rope was previously connected to a ball through a slot in the ball. When the rope is pulled through the ball, it changes location and is no longer connected to the ball. \\
\hline
KD & Rope is a string. before: ball, after: rope stroke \\
\hline
SeqKD & Rope is made of rope. \\
\hline
GKD & before: separate from Ball, after: through slot In Ball, attr: location \\
\hline
MiniLLM & through slot in ball \\
\hline
PromptKD & Rope was separate from Ball until through slot in Ball. \\
\hline
\end{tabular}
\caption{Sample response from the S-NI dataset. PromptKD follows the instructions most accurately.}
\label{tab:quant}
\end{table*}
% Appendix~\ref{sec:appendix} for an example.

\begin{table*}[!t]
\centering
\begin{tabular}{l|p{0.8\linewidth}}
\hline
\multicolumn{2}{c}{Case 1} \\
\hline
Instruction & You are given two passages. For each word in the first passage, you need to find its synonym from the second passage such that both words have the same meaning and belong to the same part of speech. The input consists of two paragraphs with a list of words following each paragraph. The output is a list of synonyms (one for each word in the first paragraph) such that they have similar meaning as corresponding words in the first paragraph and belong to same parts of speech. Note: Some sentences may not contain a particular type of word \\
\hline
Input & Paragraph 1: They water their crops with canal water. Paragraph 2: They irrigate their plants with pipeline fluid. \\
\hline
Ground Truth & Canal water and pipeline fluid are both synonyms for irrigation. \\
\hline
\hline
MiniLLM & Paragraph 2 synonyms for Paragraph 1: They irrigate Their plants with pipeline fluid. \\
\hline
PromptKD & Canal water, pipeline fluid \\
\hline
\hline
\multicolumn{2}{c}{Case 2} \\
\hline
Instruction & You are given a list of numerical values. Find the value that is halfway between the maximum and minimum values in the list, and output that value. \\
\hline
Input & List = [1, 3, 4, 6, 8]. \\
\hline
Ground Truth & The value halfway between the maximum and minimum values in the list is 4. \\
\hline
\hline
MiniLLM & The value that is halfwaybetween the maximum and minimum values is 4. This is because 4 is the average of the list, which is 1 + 3 + 4 + 6 + 8 / 5 = 4. \\
\hline
PromptKD & The median value is 4. \\
\hline
\end{tabular}
\caption{Sample response from the UnNI dataset. The Llama model is used here. Similar to the results with GPT-2, PromptKD generates appropriate responses following the instructions better than MiniLLM.}
\label{tab:quant_llama}
\end{table*}

\begin{table}[!t]
\centering
\begin{tabular}{lccc}
\hline
Method & \multirow{2}{*}{ROUGE-L} & \multirow{2}{*}{\#Params} & Time \\
(rank/alpha) & & & (hour) \\
\hline
\hline
Prompt tuning & \textbf{22.25} & \textbf{11.2K} & \textbf{9.78} \\
RoLA (4/32) & 21.84 & 1.229M & 9.97 \\
RoLA (8/16) & 21.66 & 2.458M & 10.13 \\
\hline
\end{tabular}
\caption{Comparison results according to the parameter-efficient fine-tuning method used for modifying the teacher.}
\label{tab:ablation_peft}
\end{table}

\section{GPT-4 Feedback Score}
\label{sec:gpt4_score}

We follow the approach described in Appendix D.1 of MiniLLM~\citep{minillm} to measure the GPT-4 feedback score. We utilize the GPT-4 model with a temperature of 0.7. To evaluate model output compared to ground truth response, we employ a fixed form of prompt consisting of instruction, input, assistant 1, and assistant 2. The instruction of task and input are entered first, followed by the model output in assistant 1 and the ground truth response in assistant 2, as shown in Table~\ref{tab:gpt4_score}. Through this prompt, scores for the model output and ground truth response, which are separated by spaces and range from 1 to 10, are obtained. The sum of the model output scores is divided by the sum of the ground truth scores to calculate the GPT-4 feedback score for each method. Similar to the main result in Table~\ref{tab:main_result}, scores are calculated for seeds 10, 20, 30, 40, and 50, then the average is taken. Following this approach, we measure the GPT-4 feedback scores for MiniLLM~\citep{minillm}, GKD~\citep{gkd}, and PromptKD, which demonstrated strong performance in Table~\ref{tab:main_result}. Here, we omit KD~\citep{sanh2019distilbert} and SeqKD~\citep{kim-rush-2016-sequence} from the measurement since they did not compete well against other baselines. 

\begin{table}[h]
\centering
\begin{tabular}{p{0.9\linewidth}}
\hline
Prompt Format \\
\hline
\#\#\# Instruction: \\ 
\{instruction\}  \\
\#\#\# Input: \\
\{input\} \\
\#\#\# Assistant 1: \\ 
\{model output\} \\
\#\#\# Assistant 2: \\ 
\{ground truth response\} \\
\\
We would like to request your feedback on the performance of two AI assistants in response to the user instruction and input displayed above. \\
Please rate the helpfulness, relevance, accuracy, and level of detail of their responses. Each assistant receives an overall score on a scale of 1 to 10, where a higher score indicates better overall performance. \\
Please first output a single line containing only two values indicating the scores for Assistant 1 and 2, respectively. The two scores are separated by a space. \\
In the subsequent line, please provide a comprehensive explanation of your evaluation, avoiding any potential bias and ensuring that the order in which the responses were presented does not affect your judgment. \\
\hline
\end{tabular}
\caption{Prompt format used for measuring GPT-4 feedback scores.}
\label{tab:gpt4_score}
\end{table}

The evaluation results when using GPT-2 XL (1.5B) as the teacher and GPT-2 Large (760M) as the student are summarized in Table~\ref{tab:gpt4_score_result}. Consistent with the trends observed in Table~\ref{tab:main_result}, PromptKD exhibits the best performance, followed by MiniLLM and then GKD. Particularly noteworthy is that PromptKD outperforms others on all datasets, further demonstrating the effectiveness of student-friendly knowledge. 

\section{Ablation Study}
\label{sec:abl_study}

\paragraph{Parameter-efficient fine-tuning methods}
To explore the effect of prompt tuning, we replace prompt tuning with LoRA~\citep{hu2022lora}, one of the most widely used parameter-efficient fine-tuning methods, to tune the teacher model GPT-2 XL. Since LoRA can start training from the same state as the base model, the regularization loss designed for initial stability in prompt tuning is not used here. We use the GPT-2 base model as a student. Table~\ref{tab:ablation_peft} summarizes the average instruction-following performance along with the number of trainable parameters and the training time.

From the cost perspective, using LoRA requires training approximately 100 to 200 times more parameters compared to prompt tuning, depending on the rank. This increase becomes more significant as the model size grows. This is because, in prompt tuning, the parameters are proportional to the prompt length and embedding dimension, whereas, in LoRA, they are proportional not only to the rank but also to the embedding dimension and the number of weight matrices. Additionally, training time is slightly longer when using LoRA, likely due to the increased number of trainable parameters. It is noteworthy that despite not using regularization loss, training with LoRA takes longer than with prompt tuning. Therefore, in terms of efficiency, prompt tuning is a better choice than LoRA.

In terms of performance, both prompt tuning and LoRA perform similarly, with prompt tuning having a slight edge on average. This demonstrates that it is possible to extract student-friendly knowledge with a relatively small number of trainable parameters. Furthermore, given the existing observation~\citep{lester-etal-2021-power} that prompt tuning becomes more effective with larger models, we can expect prompt tuning to be more suitable than LoRA for LLMs. Thus, using prompt tuning instead of LoRA is both an efficient and effective choice.

\begin{table}[!t]
\centering
\begin{tabular}{ccc}
\hline
\#Params & w/o $\mathcal{L}_{\text{reg}}$ & w/ $\mathcal{L}_{\text{reg}}$ \\
\hline
\hline
120M & 21.97 & \textbf{22.25} \\
340M & \textbf{24.13} & 23.92 \\
760M & 24.47 & \textbf{25.08} \\
\hline
\end{tabular}
\caption{Ablation on regularization loss. We assess the average instruction-following performance of student models without and with regularization loss to verify the effectiveness of regularization.}
\label{tab:reg_loss}
\vspace{-0.1in}
\end{table}

\paragraph{Regularization loss}

To confirm the effectiveness of the introduced regularization loss in alleviating instability when the prompt is prepended, we conduct experiments by excluding this objective. The average performance across the 5 datasets is reported in Table~\ref{tab:reg_loss}. Although there is a slight performance drop when using regularization loss with GPT-2 Medium, we observe a more significant performance increase with the other two models. This suggests the necessity of regularization loss for improving performance.
% Prompt가 prepend되었을 때의 불안정성을 해소하기 위해 도입한 regularization loss가 유효한지를 confirm하기 위해 이 objective를 제외하여 실험한 뒤 5가지 dataset의 평균 성능을 Table에 report했다. 모든 student model에서 stability loss를 제거했을 때 성능 역시 떨어지는 것을 봤을 때, regularization loss의 필요성을 확인했다.  

\paragraph{Prompt settings}

\begin{figure}[!t]
\centering
\includegraphics[width=\linewidth]{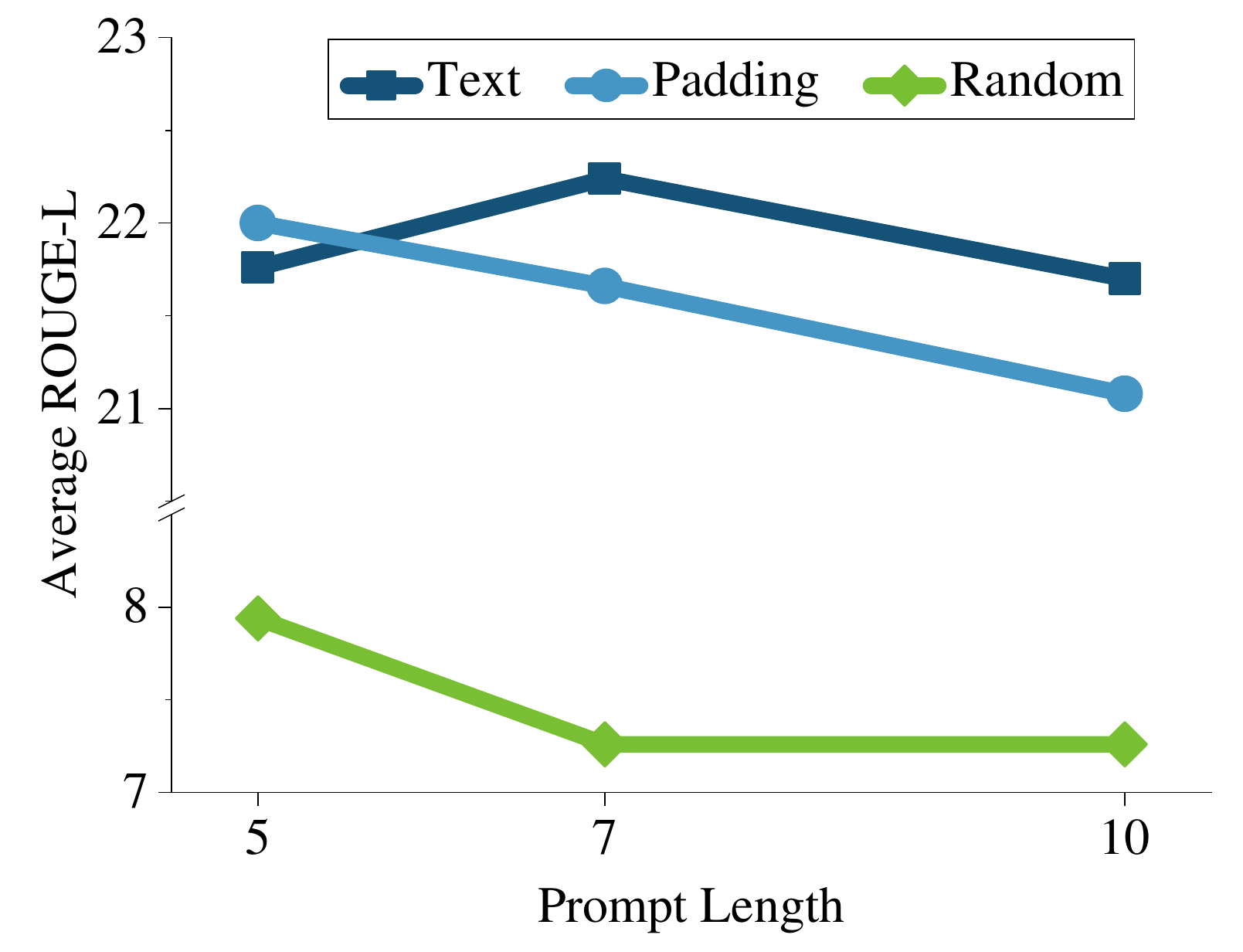}
\caption{Ablation on prompt settings. To validate the impact of prompt initialization method and length, we evaluate the average ROUGE-L score over varying these settings. }
\label{fig:prompt_length}
\end{figure}

Although the regularization loss effectively mitigates the initial instability, the prompt's length and initialization also significantly influence the prompt tuning process~\citep{hou-etal-2022-metaprompting}. Therefore, the average instruction-following performance is measured by varying the prompt length $m$ from {5, 7, 10} and the initialization method from {random, padding, text}. GPT-2 Large (760M) and GPT-2 XL (1.5B) are utilized for this ablation study. Results are summarized in Figure~\ref{fig:prompt_length}. In the padding method, all prompt tokens are initialized with the embedding of the "[PAD]" token, while in the text method, the sentence "Suppose you are a student." is tokenized, and these embeddings are used for initializing prompt tokens from the beginning. In this case, if the number of prompt tokens is smaller, the sentence is truncated, while if it is larger, all embeddings of the sentence are assigned, and then the embeddings are assigned again from the beginning for the next prompt token. 
% 비록 regularization loss를 통해 prompt tuning 과정의 초기 불안정성을 해소하였지만, prompt의 길이와 initialization 역시 prompt tuning 과정에 큰 영향을 준다 (citation). 따라서 prompt length l을 {5, 7, 10}에서, initialization 방법을 {random, padding, text}에서 조합하여 측정한 성능을 Figure에 정리했다. padding의 경우 모든 prompt token이 pad token의 embedding으로 초기화되며, text의 경우 “Suppose you are a student.” 문장을 tokenize한 뒤 각 token의 embedding으로 앞에서부터 초기화된다. 이때 prompt token의 개수가 더 작다면 문장이 잘리게 되며, 더 크다면 문장의 처음 부분부터 다시 반영한다. 

Firstly, considering the emphasis on the importance of prompt initialization in previous works, it is found that training does not proceed properly with random initialization. Moreover, generally, the text initialization method shows better performance than the padding method. Regarding length, when initialized with text, better performance is observed with a length of 7, while with padding initialization, shorter lengths exhibit better performance. This is presumably because, in text initialization, the sentence is fully encoded since it is tokenized into 7 tokens, while in padding initialization, longer lengths exert a greater influence on the instability of teacher model distribution when prepended. Therefore, all experiments in this paper are performed with a prompt length of 7, initialized using text initialization.
% 먼저, previous works에서 prompt initialization의 중요성이 강조된 만큼, random initialization을 진행하면 학습이 제대로 진행되지 않았다. 또한, 대체로 pad token보다 text로 초기화되었을 때 더욱 좋은 성능을 보였다. Length의 경우 text initialization일때는 length가 7일때, pad initialization일때는 length가 작을수록 좋은 성능을 보였다. 이는 text 초기화의 경우 해당 문장이 tokenize되었을 때 7 token이어서 가장 온전히 문장이 들어갔기 때문으로 추측되며, PAD 초기화의 경우 길이가 길어질수록 prepend될때 teacher model distribution에 더욱 큰 영향을 끼치기 때문으로 추측된다. 이 논문에서 진행된 모든 실험은 length가 7인 propmt를 text initialization하여 진행했다.

\paragraph{KL divergences}

\begin{table}[!t]
\centering
\begin{tabular}{cc}
\hline
$\mathcal{L}_{\text{kd}}$ \& $\mathcal{L}_{\text{reg}}$ & ROUGE-L \\ 
\hline
\hline
Reverse KL \& Reverse KL & \textbf{22.25} \\
Reverse KL \& Forward KL & 21.91 \\
Forward KL \& Reverse KL & 22.20 \\
Forward KL \& Forward KL & 22.13 \\
\hline
\end{tabular}
\caption{Ablation on distribution discrepancy metric. Since each loss can compute distribution discrepancy with either forward or reverse, we report the average instruction-following performance for each pair.}
\label{tab:kl_divergence}
\end{table}

To assess the impact of distribution discrepancy metrics, we conduct an ablation study on this with the same model setting. 
%We observe the best performance when measuring discrepancy using reverse KL divergence in both KD loss $\mathcal{L}_{\text{kd}}$ and regularization loss $\mathcal{L}_{\text{reg}}$, similar to previous observations. Detailed experimental results are provided in the Appendix~\ref{sec:kl_div}.
During prompt tuning, PromptKD minimizes the reverse KL divergence between the teacher distribution and the student distribution ($\mathcal{L}_{\text{kd}}$) or between the teacher distribution and the teacher distribution excluding the prompt ($\mathcal{L}_{\text{reg}}$). In this context, forward KL divergence can also be considered instead of reverse KL divergence. As shown in Table~\ref{tab:kl_divergence}, experimental results indicate that using reverse KL divergence yields the best performance. However, there is barely any significant difference. We conjecture that since the model distribution being trained is derived from the teacher, resulting in similar or even more modes in distribution, which prevent undesirable behaviors such as mode-covering even during forward KL divergence minimization.

\end{document}